\documentclass[9pt]{article}

\usepackage{lipsum} \usepackage[version=4]{mhchem}
\usepackage{siunitx}
\DeclareSIUnit\Molar{M}

\usepackage{amsmath}
\usepackage{mathtools}
\usepackage{bm}
\usepackage{float}
\usepackage{subcaption}

\usepackage{authblk}
\usepackage{natbib}
\usepackage{url}
\usepackage{booktabs}

\RequirePackage{graphicx,xcolor}
\definecolor{DarkGrey}{HTML}{273B81}
\definecolor{LightGrey}{HTML}{0A9DD9}
\definecolor{MediumGrey}{HTML}{6D6E70}
\definecolor{LightGrey}{HTML}{929497}
\definecolor{grey}{HTML}{000000}

\usepackage{algorithm}
\usepackage[noend]{algpseudocode}
\algnewcommand\algorithmicinput{\textbf{Input:}}
\algnewcommand\algorithmicoutput{\textbf{Output:}}
\algnewcommand\Input{\item[\algorithmicinput]}\algnewcommand\Output{\item[\algorithmicoutput]}
\DeclareMathOperator*{\argmax}{arg\,max}
\DeclareMathOperator*{\argmin}{arg\,min}
\newcommand{\dbf}{\mathbf{d}}
\newcommand{\ybf}{\mathbf{y}}
\newcommand{\thetab}{\bm{\theta}}
\newcommand{\psib}{\bm{\psi}}

\usepackage{mdframed}
\newcounter{textbox}
\newenvironment{textbox}[1]{  \refstepcounter{textbox}     \begin{mdframed}[hidealllines=true,backgroundcolor=grey!10,fontcolor=grey,leftline=false,linecolor=grey,linewidth=1em]
  \noindent{\bfseries\large\color{grey}Box \arabic{textbox}. #1\par}
  \if@reqslineno\addtolength{\linenumbersep}{2em}\internallinenumbers\fi
}{  \end{mdframed}
}

\newcounter{appendix}
\newenvironment{appendixbox}{  \setcounter{figure}{0}
  \setcounter{table}{0}
  \refstepcounter{appendix}  \clearpage    \begin{mdframed}[hidealllines=true,backgroundcolor=grey!10,fontcolor=grey,leftline=false,linecolor=grey,linewidth=1em]
  \noindent{\bfseries\Large\color{grey}Appendix Box \arabic{appendix}.\par}
  \if@reqslineno\addtolength{\linenumbersep}{2em}\internallinenumbers\fi
}{  \end{mdframed}
}

\title{Designing Optimal Behavioral Experiments Using Machine Learning}

\author[1\thanks{These authors contributed equally to this work. \newline Correspondence: s.valentin@ed.ac.uk}]{Simon Valentin}
\author[1$^*$]{Steven Kleinegesse}
\author[2]{Neil R. Bramley}
\author[1]{Peggy Seriès}
\author[1]{Michael U. Gutmann}
\author[1]{Christopher G. Lucas}

\affil[1]{School of Informatics, University of Edinburgh, UK}
\affil[2]{Department of Psychology, University of Edinburgh, UK}

\begin{document}

\maketitle

\begin{abstract}
Computational models are powerful tools for understanding human cognition and behavior. They let us express our theories clearly and precisely, and offer predictions that can be subtle and often counter-intuitive. However, this same richness and ability to surprise means our scientific intuitions and traditional tools are ill-suited to designing experiments to test and compare these models.
To avoid these pitfalls and realize the full potential of computational modeling, we require tools to design experiments that provide clear answers about what models explain human behavior and the auxiliary assumptions those models must make.
Bayesian optimal experimental design (BOED) formalizes the search for optimal experimental designs by identifying experiments that are expected to yield informative data.
In this work, we provide a tutorial on leveraging recent advances in BOED and machine learning to find optimal experiments for any kind of model that we can simulate data from, and show how by-products of this procedure allow for quick and straightforward evaluation of models and their parameters against real experimental data.
As a case study, we consider theories of how people balance exploration and exploitation in multi-armed bandit decision-making tasks. 
We validate the presented approach using simulations and a real-world experiment. As compared to experimental designs commonly used in the literature, we show that our optimal designs more efficiently determine which of a set of models best account for individual human behavior, and more efficiently characterize behavior given a preferred model. 
At the same time, formalizing a scientific question such that it can be adequately addressed with BOED can be challenging, and we discuss several potential caveats and pitfalls that practitioners should be aware of.
We provide code to replicate all analyses as well as tutorial notebooks and pointers to adapt the methodology to different experimental settings.
\end{abstract}

\section{Introduction}
Computational modeling of behavioral phenomena is currently experiencing rapid growth, in particular with respect to methodological improvements. 
For instance, seminal work by~\citet{wilson2019ten} has been crucial in raising methodological standards and bringing attention to how computational analyses can add value to the study of human (and animal) behavior. 
Meanwhile, in most instances, computational analyses are applied to data collected from experiments that were designed based on intuition and convention. That is, while experimental designs are usually motivated by scientific questions in mind, they are often chosen without explicitly and quantitatively considering how informative these data might be for the computational analyses and, finally, the scientific questions being studied.
This can, in the worst case, completely undermine the research effort, especially as particularly valuable experimental designs can be counter-intuitive. 
Today, advancements in machine learning open up the possibility of applying computational methods when deciding how experiments should be designed in the first place, to yield data that are maximally informative with respect to the scientific question at hand. 

In this work, we provide an introduction to modern BOED and a step-by-step tutorial on how these advancements can be combined and leveraged to find optimal experimental designs for any computational model that we can simulate data from, and show how by-products of this procedure allow for quick and straightforward evaluation of models and their parameters against real experimental data.
Formalizing the scientific goal of an experiment such that our design aligns with our intention can be difficult. 
BOED forces the researcher to make explicit, and engage with, various assumptions and constraints that would otherwise remain implicit, which helps in making experimental research more rigorous but can also lead to unexpected optimal designs. 
We discuss several issues related to experimental design that practitioners may encounter no matter how they design their experiments. Importantly, these issues are made more salient by formalizing the process. 

\section{Why optimize experimental designs?}
Experiments are the bedrock of scientific data collection, making it possible to discriminate between different theories or models, and discovering what specific commitments a model must make to align with reality by means of parameter estimation.
The process of designing experiments involves making a set of complex decisions, where the space of possible experimental designs is typically navigated based on a combination of researchers' prior experience, intuitions about what designs may be informative, and convention.

This approach has been successful for centuries across various scientific fields, but it has important limitations that are amplified as theories become richer, more nuanced and more complex. Often, a theory does not make a concrete prediction, but rather has free parameters that license a variety of possible predictions of varying plausibility. 
For example, if we want to model human behavior, it is important to recognize that different participants in a task might have different strategies or priorities, which can be captured by parameters in a model. As we expand models to accommodate more complex effects, it becomes ever more time-consuming and difficult to design experiments that distinguish between models, or allow for effective parameter estimation. 
As an example from cognitive science, there are instances where models turned out to be empirically indistinguishable under certain conditions, a fact that was only recognized after a large number of experiments had already been done~\citep{jones2014unfalsifiability}. Such problems also play an important role in the larger context of the replication crisis~\citep[e.g.,][]{pashler2012}. 

Collecting experimental data is typically a costly and time consuming process, especially in the case of studies involving resource-intensive recording techniques, such as neuroimaging or eye-tracking~\citep[e.g.,][]{dale1999optimal, lorenz2016automatic}. There is a strong economic as well as ethical case for conducting experiments that are expected to lead to maximally informative data with respect to the scientific question at hand.
For instance, in computational psychiatry, researchers are often interested in estimating parameters describing people's traits and how they relate, often doing so under constrained resources in terms of the number of participants and their time. Additionally, researchers and practitioners alike require that experiments are sufficiently informative to provide confidence that they would reveal effects if they are there and also that negative results are true negatives~\citep[e.g.,][]{huys2013mapping, karvelis2018autistic}.
Consequently, as theories of natural phenomena become more realistic and complex, researchers must expand their methodological toolkit for designing scientific experiments.

\subsection{Bayesian optimal experimental design}
As a potential solution, \emph{Bayesian optimal experimental design}~\citep[BOED; see][for reviews]{ryan_review_2016, rainforth2023modern} provides a principled framework for optimizing the design of experiments. 
Broadly speaking, BOED rephrases the task of finding optimal experimental designs to that of solving an optimization problem.
That is, the researcher specifies all controllable parameters of an experiment, also known as experimental designs, for which they wish to find optimal settings, which are then determined by maximizing a utility function.
Exactly what we might want to optimize depends on our experiment, but can include any aspect of the design that we can specify, such as which stimuli to present, when or where measurements should be taken in a quasi-experiment, or, as an example, how to reward risky choices in a behavioral experiment.
The utility function is selected to measure the quality of a given experimental design with respect to the scientific goal at hand, such as model discrimination, parameter estimation or prediction of future observations, with typical choices including expected information gain, uncertainty reduction, and many others~\citep[][]{ryan_review_2016}, as explained below.
Importantly, BOED approaches require that theories are formalized via computational models of the underlying natural phenomena.

\section{Computational models}
Scientific theories are increasingly formalized via computational models~\citep[e.g.,][]{guest2021computational}. These models take many forms, but one desideratum for a model is that, given a set of experimental data, we can compare it to other models based on how well it predicts that data. 
To that end, many computational models are constructed in a way that allows for the analytic evaluation of the likelihood of the model given data, by means of a likelihood function.
However, this imposes strong constraints on the space of models we can consider. Models that are rich enough to accurately describe complex phenomena---like human behavior---often have intractable likelihood functions~\citep[e.g.,][]{lintusaari_fundamentals_2017, cranmer_frontier_2020}. In other words, we cannot compute likelihoods due to the prohibitive computational cost of doing so, or the inability to express the likelihood function mathematically. 
This has at least two implications (1) When we develop rich models, we often lack the tools to evaluate and compare them directly, and are left with coarse, qualitative proxy measures--like the ``the model curves look like the human curves for some parameter settings!''; (2) when we are intent on conducting a systematic comparison of models, we are often forced to use simplified models that have tractable likelihoods, even when the simplifications lead to consequential departures from the theories we might want our models to capture.

\subsection{Simulator models: What if the likelihood cannot be computed?}
Many realistically complex models have the feature that we can \emph{simulate} data from them, which allows for using simulation-based inference methods, such as approximate Bayesian computation (ABC) and many others~\citep[e.g.,][]{lintusaari_fundamentals_2017, cranmer_frontier_2020}.
There has been a growing interest in this class of models, often referred to as \emph{generative}, \emph{implicit} or \emph{simulator} models~\citep[e.g.,][]{marjoram2003markov, sisson2007sequential, lintusaari_fundamentals_2017, brehmer2020mining, cranmer_frontier_2020}.
We will use the term simulator model throughout this paper to refer to any model from which we can simulate data. We note that some generative models have tractable likelihood functions, but this class of models is subsumed by simulator models, which do not make any structural assumptions on the likelihood and include important cases where the likelihood is too expensive to be computed, or cannot be computed at all.\footnote{Simulator models also include the common case of a potentially complex deterministic simulator with an additional observation noise component, as is used in many areas of science.}
Simulator models have become ubiquitous in the sciences, including physics~\citep{schafer2012}, biology~\citep{ross2017}, economics~\citep{gourieroux2010}, epidemiology~\citep{corander2017} and cognitive science~\citep{palestro2018likelihood}, among others.
\begin{figure*}[t!]
    \centering
    \includegraphics[width=1\linewidth]{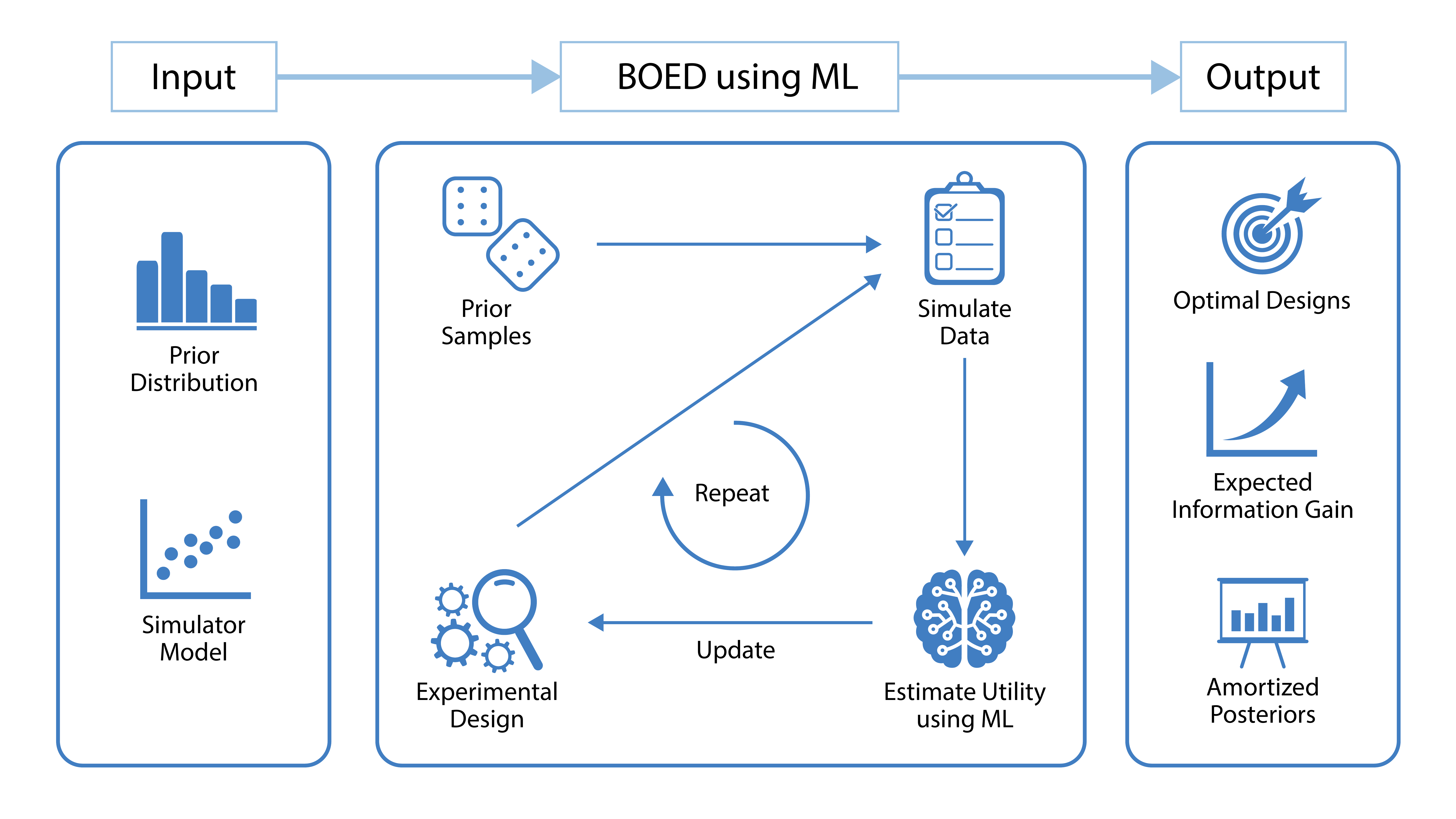}
    \vspace{-0.65cm}
    \caption{A high level overview of our approach that uses machine learning (ML) to optimize the design of experiments. The inputs to our method are a model from which we can simulate data and a prior distribution over the model parameters. Our method starts by drawing a set of samples from the prior and initializing an experimental design. These are used to simulate artificial data using the simulator model. We then use ML to estimate the expected information gain of those data, which is used as a metric to search over the design space. Finally, this is repeated until convergence of the expected information gain. The outputs of our method are optimal experimental designs, an estimate of the expected information gain when performing the experiment and an amortized posterior distribution, which can be used to cheaply compute approximate posterior distributions once real-world data are observed. Other useful by-products of our method include a set of approximate sufficient summary statistics and a fitted utility surface of the expected information gain.}
    \label{fig:main_schematic}
\end{figure*}

The study of human, and animal, perception and cognition is experiencing a surge in the use of computational modeling~\citep[e.g.,][]{griffiths2015manifesto}.
Here, simulator models can formalize complex theories about latent psychological processes to arrive at quantitative predictions about observable behavior.
As such, simulator models appear across cognitive science, including e.g., many Bayesian and connectionist models, a wide range of process models, cognitive architectures, or many reinforcement learning models and physics engines~\citep[see, e.g.,][]{turner2012tutorial, turner2014generalized, bitzer2014perceptual, turner2018competing, bramley2018intuitive, ullman2018learning, palestro2018likelihood, kangasraasio_parameter_2019, gebhardt_hierarchical_2020}.
The applicability and usefulness of experimental design optimization for parameter estimation and model comparison has been demonstrated in many scientific disciplines, but this has often been restricted to settings with simple and tractable classes of models, for instance in cognitive science~\cite[e.g.,][]{myung_optimal_2009, zhang_optimal_2010, ouyang_webppl-oed_2018}. 

\section{An overview of this tutorial}
In this tutorial, we present a flexible workflow that combines recent advances in machine learning (ML) and Bayesian optimal experimental design (BOED) in order to optimize experimental designs for any computational model from which we can simulate data. A high-level overview of the underlying ML-based BOED method is shown in Figure~\ref{fig:main_schematic}, including researchers' inputs and resulting outputs.
The presented approach is applicable to any model from which we can simulate data and scales well to realistic numbers of design variables, experimental trials and blocks. 
Furthermore, as well as optimized designs, the presented approach also results in tractable amortized posterior inference---that is, it allows researchers to use their actual experimental data to easily compute posterior distributions, which may otherwise be computationally expensive or intractable to compute.
Moreover, we note that, to our knowledge, there has been no study that performs BOED for simulator models without likelihoods and then uses the optimal designs to run a real experiment. 
Instead, most practical applications of BOED have been limited to simple scientific theories of nature~\citep[e.g.,][]{liepe2013}.
This paucity of real-world applications is partly because the situations where BOED would be most useful are those where it has previously been computationally intractable to use. 
The ML-based approach presented in this work aims to overcome these limitations.

As with many new computational tools, however, the techniques for using BOED are not trivial to implement from scratch. In an effort to help other scientists understand these methods and use them to unlock the full potential of BOED in their own research, we next present a case study in using BOED to better understand human decision-making. Specifically, we (1) describe new simulator models that generalize earlier decision-making models that were constrained by the need to have tractable likelihoods; (2) walk through the steps to use BOED to design experiments comparing these simulator models and estimating the psychologically interpretable parameters they contain; and (3) contrast our new, optimized experiments to previous designs and illustrate their advantages and the new lessons they offer. All of our steps and analyses are accompanied by code that we have commented and structured to be easily adapted by other researchers.

\section{Case study: Multi-armed bandit tasks}
As a case study application, we consider the question of how people balance their pursuit of short-term reward (``exploitation'') against learning how to maximize reward in the longer term (``exploration''). This question has been studied at length in psychology, neuroscience, and computer science, and one of the key frameworks for investigating it is the multi-armed bandit decision-making task~\cite[e.g.,][]{steyvers_bayesian_2009, dezfouli2019models, schulz2020finding,gershman2018deconstructing}.
Multi-armed bandits have a long history in statistics and machine learning~\cite[e.g.,][]{robbins_aspects_1952, sutton_reinforcement_2018} and formalize a general class of sequential decision problems---repeatedly choosing between a set of options under uncertainty with the goal of maximizing cumulative reward (for more background, see Box~\ref{box:bandit_tasks}).

We demonstrate the applicability of our method by optimizing the design of a multi-armed bandit decision-making task, considering three flexible extensions of previously proposed models of human behavior.
In our experiments, we consider two general scientific tasks: model discrimination and parameter estimation.
We then evaluate the optimal designs found using our method empirically with simulations and real human behavioral data collected from online experiments, and compare them to designs commonly used in the literature.
We find that, as compared to commonly-used designs, our optimal experimental designs yield significantly better model recovery, more informative posterior distributions and improved model parameter disentanglement. 

\begin{textbox}{Bandit tasks}
\label{box:bandit_tasks}
Bandit tasks constitute a minimal reinforcement learning task, in which the agent is faced with the problem of balancing exploration and exploitation.
Various algorithms have been proposed for maximizing the reward in bandit tasks, and some of these algorithms have been used directly or in modified form as models of human behavior. 
A selection of such algorithms indeed captures important psychological mechanisms in solving bandit tasks~\citep{steyvers_bayesian_2009}. 
However, a richer account of human behavior in bandit tasks would seem to be required to accommodate flexibility and nuance of human behavior~\cite[e.g., see][]{dezfouli2019models, lee_psychological_2011}, leading to simulator models that do not necessarily permit familiar likelihood-based inference, due to, e.g., unobserved latent states, and which complicate the task of intuiting informative experiments. See Figure~\ref{fig:simulator_sketch} for a schematic of the data-generating process of multi-armed bandit tasks.

\begin{center}
    \includegraphics[width=0.8\linewidth]{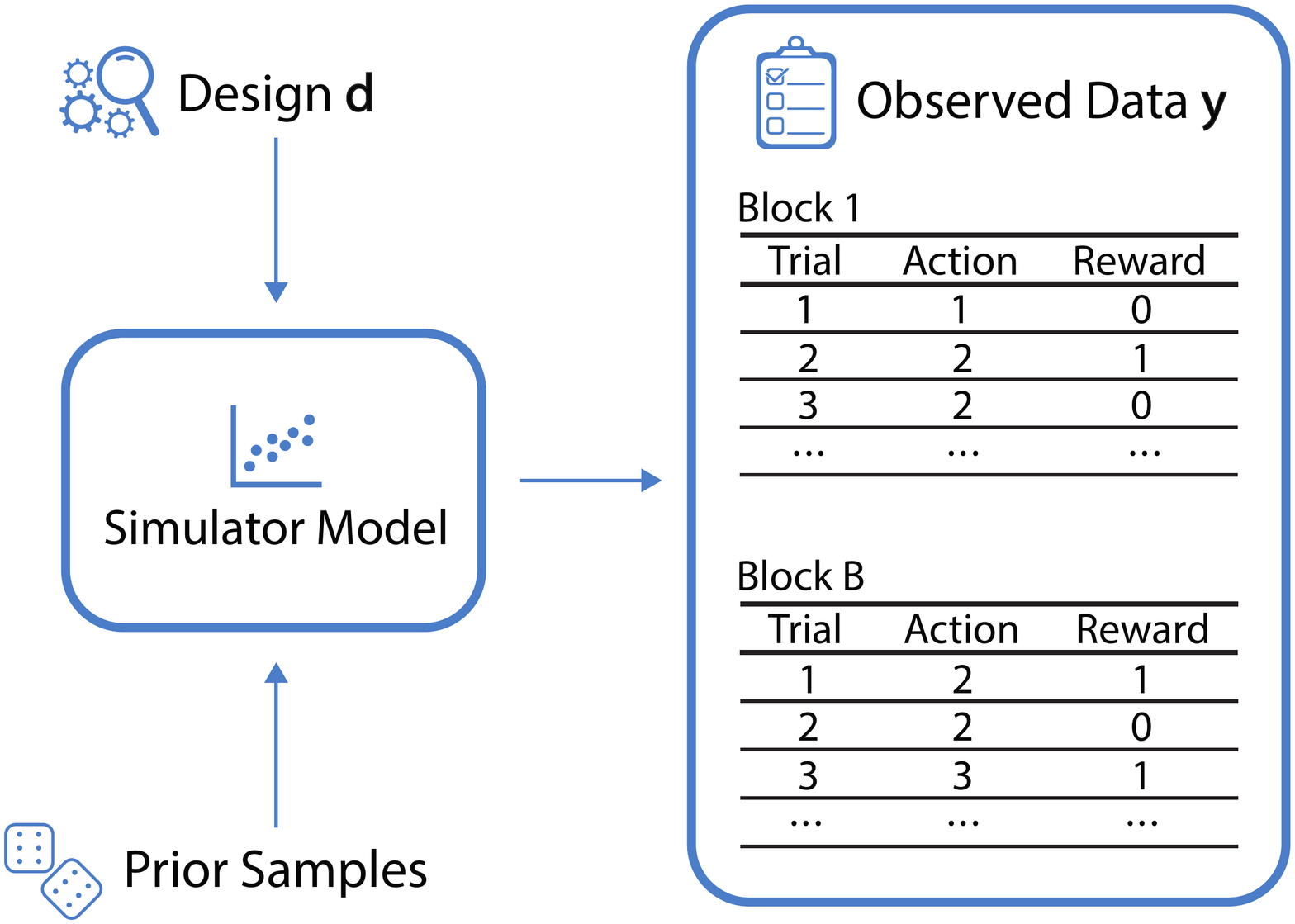}
    \captionof{figure}{Schematic of the data-generating process for the example of multi-armed bandit tasks. For a given design $\bm{d}$ (here the reward probabilities associated with the bandit arms in each individual block) and a sample from the prior over the model parameters $\bm{\theta}$, the simulator generates observed data $\bm{y}$, corresponding to the actions and rewards from $B$ blocks. }
    \label{fig:simulator_sketch}
\end{center}
\end{textbox}

\section{Workflow Summary}
In the following sections, we describe and walk through our workflow step-by-step, using the aforementioned multi-armed bandit task setting as a concrete case study to further illustrate ML-based BOED.
At a high level, our workflow comprises the following steps: 
\begin{enumerate}
    \itemsep0.5em
    \item Defining a scientific goal, e.g., which model, among a set of models, best describes a natural phenomenon.
    \item Formalizing a theory, or theories, as a computational model(s) that can be sampled from.
    \item Setting up the design optimization problem and deciding which aspects of the experimental design need to be optimized.
    \item Constructing the required machine learning models and training them with simulated data.
    \item Validating the obtained optimal designs in silico and, potentially, re-evaluating design and modeling choices.
    \item After confirming and validating the optimal design, the real experiment can be performed. The trained machine learning models used for BOED can be used afterwards to easily compute posterior distributions.  
\end{enumerate}

We provide a more detailed description of the above workflow in the following sections. The outputs of the ML-based BOED method are optimal experimental designs, an estimate of the (maximum) expected information gain and an amortized posterior distribution, which allows for straightforward computation of posterior distributions from collections of real-world data and saves us a potentially computationally expensive (likelihood-free) inference step. 
Readers may find that many of these steps should be part of the design of experiments no matter whether using BOED or not. However, BOED here provides the additional advantage of making all of these steps, which are sometimes carried out heuristically, explicit and formal.

\section{Step 1: Formulate your scientific goal}
The first step in setting up the experimental design optimization is to define our scientific goal. 
While there may be many different goals for an experiment, they often fall into the broad categories of model discrimination or parameter estimation.  

\subsection{Model discrimination} Computational models provide formal and testable implementations of theories about nature. Model discrimination (also often referred to as \emph{model comparison}) lies at the core of many scientific questions and amounts to the problem of deciding which of a set of different models best explains some observed data. 
A rigorous formulation of this problem is given by the Bayesian approach: 
Starting from a prior belief about the plausibility of different models and their parameters, we consider how well different models can explain the observed data and use this information to update our prior beliefs. Crucially, more flexible theories are penalized automatically via the notion of the Bayesian Ockham's razor. This ensures that in cases where two theories can explain the data equally well, we should favor the simpler one. 
Formally, this requires computing the marginal likelihood of our different models, which may mean integrating (or for discrete parameters, summing) over a large number of model parameters, which is often intractable, and thus often tackled via information criteria, which may, however, suffer from various problems~\citep[e.g.,][]{pitt2002good, dziak2020sensitivity, schad2021toward}. 
This computational complexity of computing marginal model likelihoods is exacerbated when model parameter likelihood is intractable, as is the case for many simulator models.

With respect to our case study, we may, for instance, be interested in comparing which of the three simulator models presented in Box~\ref{box:2} best explains human behavior in multi-armed bandit tasks.

\subsection{Parameter estimation}
Researchers are often interested in learning about model parameters of a particular, single model. Similar to the model discrimination approach, a Bayesian approach of learning about these parameters would involve starting with a prior distribution and then updating our belief via Bayes' rule using observed data. For instance, we may be interested in estimating an individual's (or population's) working memory capacity given a working memory model, response time distribution in a model for selective attention, choice probability of choosing between different options in a risky choice model, inclination to explore versus exploit in a reinforcement task, or associations between biases in probabilistic visual integration tasks and autistic/schizotypal traits~\citep{karvelis2018autistic} to name a few examples. 
Model parameters may here be continuous or discrete, which matters for how we set up up the optimization procedure in Step 4.

As an example, for the Auto-regressive $\epsilon$-Greedy (AEG) simulator model in our case study, we are, among many other things, interested in estimating peoples' inclination to re-select previously chosen options as opposed to avoiding repeating their own past behavior, which is characterized by a specific model parameter. 

\subsection{Other goals}
While we here focus on the problems of model discrimination and parameter estimation, we point out that the overall BOED approach via mutual information estimation is more general, and can easily be adapted to other tasks, such as improving future predictions~\citep[e.g.][]{kleinegesse2021_gradbed} and many others~\citep[see][]{ryan_review_2016}.

\subsection{How do we measure the value of an experimental design?}
We formalize the quality of our experimental design via a utility function, which thereby serves as a quantitative way of measuring the value of one experimental design over any another and provides the objective function for our optimization problem.
Following recent work in BOED~\citep[e.g.,][]{kleinegesse2019, kleinegesse_bayesian_2020, kleinegesse2020sequential, kleinegesse2021_gradbed, foster_variational_2019}, we use \emph{mutual information}~\citep[MI;][]{lindley1952} (also known as the \emph{expected information gain}) as our utility function $U(\mathbf{d})$ for measuring the value of an experimental design $\dbf$, i.e.
\begin{equation} \label{eq:mi}
U(\mathbf{d})=E_{p(\mathbf{y} \mid \boldsymbol{v}, \mathbf{d}) p(\boldsymbol{v})}\left[\log \frac{p(\boldsymbol{v} \mid \mathbf{y}, \mathbf{d})}{p(\boldsymbol{v})}\right],
\end{equation}
where $\bm{v}$ is a variable we wish to estimate and $\bm{y}$ corresponds to the observed data (actions and rewards in our case study). Note that the probability density function $p(\bm{v}|\mathbf{y}, \mathbf{d})$ is the posterior distribution and $p(\bm{v})$ is the prior distribution of the variable of interest.

For the model discrimination task, the variable of interest $\bm{v}$ in our utility function corresponds to a scalar model indicator variable $m$ (where $m=1$ corresponds to a model 1, $m=2$ to some model 2 and so on for as many models as we are comparing).
For the parameter estimation task, the variable of interest $\bm{v}$ in our utility function is the vector of model parameters $\thetab_m$ for a given model $m$. 

Mutual information has several appealing properties~\citep{paninski2005} that make it a popular utility function in BOED~\citep{ryan_review_2016}. Intuitively, MI quantifies the amount of information our experiment is expected to provide about the variable of interest. Furthermore, the MI utility function can easily be adapted to various scientific goals by changing the variable of interest~\citep{kleinegesse2021_gradbed}. Optimal designs are then found by maximizing this utility function, i.e.~$\mathbf{d}^\ast = \argmax U(\mathbf{d})$. Although a useful quantity, estimating $U(\mathbf{d})$ is typically extremely difficult, especially for simulator models that have intractable likelihoods $p(\mathbf{y} | \bm{v}, \mathbf{d})$~\citep[e.g.,][]{ryan_review_2016}. Nonetheless, in the later steps of our workflow we shall explain how we can approximate this utility function using ML-based approaches.

\section{Step 2: Cast your theory as a computational model}
After defining our scientific goal, we need to ensure that our scientific theories are, or can be, formalized as computational models. This either requires us to formalize them from the ground up, or leverage existing computational models from literature that support our theories. In the former case, this entails making all aspects of the theory explicit, which is therefore a highly useful and established practice; we refer the reader to~\cite{wilson2019ten} for a more general guidance on building computational models. For the presented ML-based BOED approach, we require that our theories are cast as computational models that allow us to simulate data, i.e. simulator models. Further we need to decide on a general experimental paradigm that needs to be able to interact with our computational models. 

\subsection{Building a computational model}
Formally, a computational model defines a generative model $\mathbf{y} \sim p(\mathbf{y} | \bm{\theta}, \mathbf{d})$ that allows sampling of synthetic data $\mathbf{y}$, given values of its model parameters $\bm{\theta}$ and the experimental design $\mathbf{d}$. We assume that sampling from this model is feasible, but make no assumptions about whether evaluating the likelihood function $p(\mathbf{y} | \bm{\theta}, \mathbf{d})$ is tractable or intractable. Even if the likelihood is tractable and simple, important quantities such as the marginal likelihood may nevertheless be intractable, since evaluating the marginal likelihood requires integrating over all free parameters. This problem is naturally exacerbated for more complex models that may have intractable likelihood functions.

See Box~\ref{box:2} for a set of simulator models from our case study. These models describe human behavior in a multi-armed bandit setting, where the observed data $\bm{y}$ corresponds to the sequences of chosen bandit arms along with their observed reward.

\subsection{What are our prior beliefs?}
In addition to formalizing our theories as simulator models, we need to complete this step by defining prior distributions over the (free) model parameters, as would be required for any analysis. We point the reader to~\cite{wilson2019ten, schad2021toward, mikkola2021prior} for guidance on specifying prior distributions, as this is a broad research topic and a detailed treatment is beyond the scope of this work. 
Generally, prior distributions should reflect what is known about the model parameters prior to running the experiment, either based on empirical data or general domain knowledge. 
This is crucial for the design of experiments, as we may otherwise spend experimental resources and effort on learning what is already known about our models.

\begin{textbox}{Models of human behavior in bandit tasks}
\label{box:2}
We study models of human choice behavior in bandit tasks to demonstrate and validate our method.
We generalize previously proposed models (see Appendix 3 for details), and consider three novel computational models for which likelihoods are intractable.
\paragraph{Win-Stay Lose-Thompson-Sample (WSLTS)} Posits that people tend to repeat choices that resulted in a reward, and otherwise to explore by selecting options with a probability proportional to their posterior probability of being the best (Thompson sampling).
This model generalizes the Win-Stay Lose-Shift model that has been considered in previous work~\citep{steyvers_bayesian_2009}, but accommodates the possibility that people might be more likely to shift to more promising alternatives than uniformly at random.
\paragraph{Auto-regressive $\varepsilon$-Greedy (AEG)} Formalizes the idea that people might greedily choose the option with the highest estimated reward in each trial with a certain probability or otherwise explore randomly, but also permits that people may be ``sticky'', i.e., having some tendency to re-select whatever option was chosen on the previous trial, or ``anti-sticky'', i.e., preferring to switch to a new option on each trial. This generalizes the $\varepsilon$-greedy model studied previously~\citep{steyvers_bayesian_2009}, further accommodating the possibility that people preferentially stick to or avoid their previous selection.
\paragraph{Generalized Latent State (GLS)} Captures the idea that people might have an internal latent exploration-exploitation state that influences their choices when encountering situations in which they are faced with an explore-exploit dilemma. Transitions between these states may depend on previously encountered rewards and the previous latent state. 
This generalizes previous latent-state models that assume people shift between states independently of previous states and observed rewards~\cite{zhang2009human}, or only once per task~\citep{lee_psychological_2011}.
As these models are novel and to illustrate the method, we do not specify strong prior beliefs about any model parameters and describe the priors distribution in Appendix 1.
\end{textbox}

\section{Step 3: Decide which aspect of your experiment to optimize}
Having formalized our scientific theories as simulator models and decided on our scientific goals, we need to think about which parts of the experiment we want to optimize. In principle, any controllable aspect of the experimental apparatus could be tuned as part of the experimental design optimization. Concretely, in order to solve an experimental design problem by means of BOED, our experimental designs $\dbf$ need to be formalized as part of the simulator model $p(\ybf | \thetab, \dbf)$. Instead of optimizing all controllable aspects of an experiment, it may then help to leverage domain knowledge and prior work to choose a sufficient experimental design set, in order to reduce the dimensionality of the design problem.

In our case study, for instance, we have chosen the reward probabilities of different bandit arms to be the experimental designs to optimize over. This choice posits that certain reward environments yield more informative data than others, in the context of model discrimination and parameter estimation. Our task is then to find the optimal reward environment that yields maximally informative data.

There may be other experimental choices that we will have to make, in particular to limit variability in the observed data, to control for confounding effects, and to reduce non-stationarity, among many other reasons. In our multi-armed bandit setting, for example, we may choose to limit the number of trials in the experiment to avoid participant fatigue, which is usually not accounted for in computational models of behavioral phenomena. Additionally, we may limit the number of available choices in a discrete choice task or impose constraints on any other aspects suggested by domain knowledge, or where we want to ensure comparability with prior research. 
For example, we may have reason to believe that certain designs are more ecologically valid than others --- we discuss such considerations in the \emph{Caveats} Section.

\subsection{Static and adaptive designs}
We consider the common case of \emph{static} Bayesian optimal experimental design, with the possibility of having  multiple blocks of trials. That is, we find a set of optimal designs prior to actually performing the experiment. In some cases, however, a scientist may wish to optimize the designs sequentially as the participant progresses through the experiment. This is called \emph{sequential}, or \emph{adaptive}, BOED~\citep[see][]{ryan_review_2016, rainforth2023modern} and has been explored previously for tractable models of human behavior~\citep{myung_tutorial_2013}, or in the domain of adaptive testing~\citep[e.g.,][]{weiss1984application}. More recently, sequential BOED has also been extended to deal with implicit models~\citep{ kleinegesse2020sequential, ivanova2021}. 
At this point, sequential BOED is an active area of research and may be a consideration for researchers in the future. 
Generally, sequential designs can be more efficient, as they allow to adapt the experiment ``on-the-fly''. However, sequential designs also make building the actual experiment more complicated, because the experiment needs to be updated in almost real-time to provide a seamless experience. Further, in some settings participants may notice that the experiment adapts to their actions, which may inadvertently alter their behavior.
As we discuss below in the \emph{Caveats} Section, the effects of model misspecification may be especially severe with sequential designs, contributing to our focus on static designs. 
For future applications, which approach to follow is likely going to be a choice that uniquely depends on the experiment and research questions at hand, while the overall workflow outlined in this tutorial will be relevant to both.

\begin{textbox}{Experimental settings in the case study}
Behavioral experiments are often set up with several experimental blocks. We here make the common assumption of exchangeability with respect to the experimental blocks in our analyses and randomize block order in all experiments involving human participants.
In our experiments, the bandit task in each experimental block has three choice options (bandit arms) and $30$ trials (sequential choices). 
For each block of trials, the data $\ybf$ consists of $30$ actions and $30$ rewards, leading to a dimensionality of $60$ per block. 
For a given block, the experimental design $\dbf$ we wish to optimize are the reward probabilities of the multi-armed bandit, which, in our setting, are three-dimensional (corresponding to independent scalar Bernoulli reward probabilities for each bandit arm). 
The generative process for models of human behavior in bandit tasks is illustrated in Box~\ref{box:2}.

We demonstrate the optimization of reward probabilities for multi-armed bandit tasks, with the scientific goals of \emph{model discrimination} and \emph{parameter estimation}. We consider five blocks of $30$ trials per participant, yielding a fairly short but still informative experiment. We divide this into two blocks for model discrimination followed by three blocks for parameter estimation, where the parameter estimation blocks are conditional on the winning model for the participant in the first two blocks. In other words, for the parameter estimation task, a different experimental design (which corresponds to a particular setting of the reward probabilities associated with each bandit arm) is presented to a participant depending on the simulator model that best described them in the first two blocks. Since we consider three-armed bandits, the design space is constituted by the possible combinations of reward probabilities. This is $6$-dimensional for model discrimination and $9$-dimensional for PE. Similarly, the data for a single block is $60$-dimensional, so the dimensionality of the data is $120$ for model discrimination and $180$ for PE.
Our method applies to any prior distribution over model parameters or the model indicator, but for demonstration purposes we assume uninformative prior distributions on all model parameters (see Appendix 1 for details).
As our baseline designs, we sample all reward probabilities from a $\text{Beta}(2,2)$ distribution, following a large and well-known experiment on bandit problems with $451$ participants~\citep{steyvers_bayesian_2009}.
More information about the experimental settings and detailed descriptions of the algorithms can be found in the Appendix.
\end{textbox}

\section{Step 4: Use machine learning to design the experiments}
As previously explained, mutual information (MI) shown in Equation~\ref{eq:mi} is an effective means of measuring the value (i.e., the \emph{utility}) of an experimental design setting $\dbf$ in BOED~\citep{paninski2005, ryan_review_2016}. Unfortunately, it is generally an expensive, or intractable, quantity to compute, especially for simulator models. The reason for this is two-fold: Firstly, the MI is defined via a high-dimensional integral. Secondly, the MI requires density evaluations of either the posterior distribution or the marginal likelihood, both of which are expensive to compute for general statistical models, and intractable for simulator models.
Recent advances in BOED for simulator models thus advocate the use of cheaper MI lower bounds instead~\citep[e.g.][]{foster_variational_2019, kleinegesse_bayesian_2020, kleinegesse2021_gradbed, ivanova2021}.

Many recent innovations in BOED rely on machine learning to accurately and efficiently estimate functions.
Machine learning is currently revolutionizing large parts of the natural sciences, with applications ranging from understanding the spread of viruses~\citep{currie2020} to discovering new molecules \citep{jumper2021highly}, forecasting climate change~\cite{runge2019inferring} and developing new theories of human decision-making~\citep{peterson2021using}.
Due to its versatility, machine learning can be integrated effectively into a wide range of scientific workflows, facilitating theory building, data modeling and analysis.
In particular, machine learning allows us to automatically discover patterns in high-dimensional data sets. 
Many recent state-of-the-art methods in natural science therefore deeply integrate machine learning into their data analysis pipelines~\citep[e.g.,][]{blei2017science}.
However, there has been little focus on applying machine learning to improve the data collection process, which ultimately determines the quality of data, and thus the efficacy of downstream analyses and inferences. 
In this work, we use machine learning methods as part of our optimization scheme, where they serve as flexible function approximators.

The step of training a good machine learning model currently requires some previous expertise with (or willingness to learn about) applying machine learning methods. 
Fortunately, this step is less experiment-specific than it may first appear. 
Many different experiments generate similar data, and the architecture we provide can deal with multiple blocks and automatically learns summary statistics, as we explain above.
Many excellent resources on the basics of machine learning exist, and a detailed treatment is beyond the scope of this tutorial. 
However, we provide pointers to several aspects in the code repository: \url{https://github.com/simonvalentin/boed-tutorial}.

\subsection{Approximating mutual information with machine learning}
In this work, we leverage the MINEBED method~\citep{kleinegesse_bayesian_2020}, which allows us to efficiently estimate the MI using machine learning. Specifically, for a particular design $\dbf$, we first simulate data from the computational models under consideration, (shown in Box~\ref{box:2}). The approach works by constructing a lower bound on the MI that is parameterized via a machine learning model. In our case study we use a neural network with an architecture customized for behavioral experiments (shown in Box~\ref{box:4}), but any other machine learning model would work in place. We then train the machine learning model by using the simulated data as input and the lower bound as an objective function, which gradually tightens the lower bound. 

\begin{textbox}{Bespoke neural network architecture for behavioral experiments} \label{box:4}
\begin{center}
    \includegraphics[width=1\linewidth]{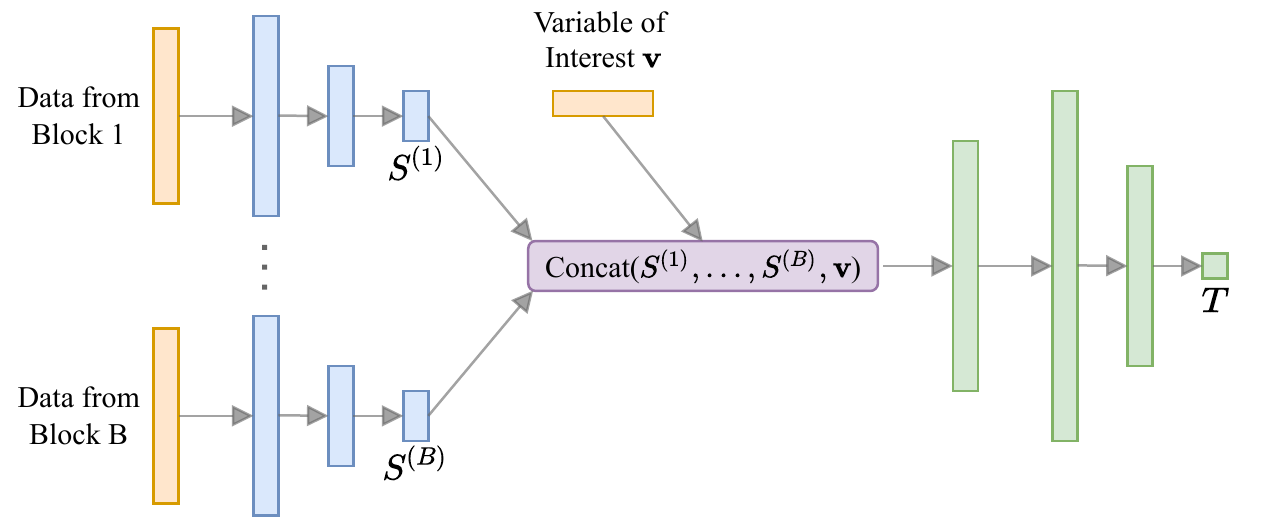}
    \captionof{figure}{Neural network architecture for behavioral experiments. For each block of data we have a small sub-network (shown in blue) that outputs summary statistics $S$. These are concatenated with the variable of interest $\bm{v}$ (e.g., corresponding to a model indicator for model discrimination tasks) and passed to a larger neural network (shown in green). Figure 3 has been adapted from Figure 1 in \citet{valentin2021bayesian}.}
    \label{fig:nn_arch}
\end{center}
\end{textbox}

More concretely, this method works by training a neural network $T_{\psib}(\bm{v}, \ybf)$, or any other machine learning model, where $\psib$ are the neural network parameters and the data $\mathbf{y}$ is simulated at design $\mathbf{d}$ with samples from the prior $p(\bm{v})$. We train this neural network by maximizing the following objective function, which is a \emph{lower bound} of the mutual information,

\begin{equation} \label{eq:lb}
U(\mathbf{d} ; \boldsymbol{\psi})=E_{p(\mathbf{y} \mid \boldsymbol{v}, \mathbf{d}) p(\boldsymbol{v})}\left[T_{\boldsymbol{\psi}}(\boldsymbol{v}, \mathbf{y})\right]-E_{p(\mathbf{y} \mid \mathbf{d}) p(\boldsymbol{v})}\left[e^{T_{\boldsymbol{\psi}}(\boldsymbol{v}, \mathbf{y})-1}\right]
\end{equation}

The above lower bound is also known as the NWJ lower bound and has several appealing bias-variance properties~\citep{poole2019}.\footnote{We note that alternative bounds on MI can be used in the place of the NWJ bound as our loss function, as discussed by~\cite{kleinegesse2021_gradbed}.} 
The expectations in Equation~\ref{eq:lb} are usually approximated using (Monte-Carlo) sample-averages.

Once we have trained the neural network $T_{\psib}(\bm{v}, \ybf)$ on $\psib$, we estimate the mutual information at $\dbf$ by computing Equation~\ref{eq:lb} using a held-out test set of data simulated from the computational model(s).

\subsection{Dealing with high-dimensional observations}
We can efficiently deal with high-dimensional data and do not have to construct (approximately sufficient) summary statistics based on domain expertise, as is commonly done~\citep{myung_optimal_2009, zhang_optimal_2010, ouyang_webppl-oed_2018}, and which can be prohibitively difficult for complex simulator models~\citep{chen2021}. In fact, the previously described method can allow us to automatically learn summary statistics as a by-product, as explained below.

Critical to the performance of our method in our case study, as with any application of neural networks, is the choice of architecture (see~\citet{elsken2019} and~\citet{ren2021} for recent reviews of the challenges and solutions in neural architecture search). In order to develop an effective neural network architecture, we leverage the method of neural approximate sufficient statistics~\citep{chen2021} and propose them as a novel and practical alternative to hand-crafted summary statistics in the context of BOED. 
Behavioral data are often collected in several blocks that can have different designs  but should be analyzed jointly, which can be challenging due to the high dimensionality of the joint data, which typically comprises numerous choices, judgments or other measurements (including high-resolution and/or high-frequency neuroimaging data). In order to deal with a multi-block data structure effectively, we propose an architecture devised specifically for application to behavioral experiments. Our architecture, visualized in Figure~\ref{fig:nn_arch} in Box~\ref{box:4}, incorporates a sub-network for each block of an experiment. The outputs of these sub-networks are then concatenated and passed as an input to a larger neural network network. In doing so, besides being able to reduce the dimensionality of the input data, each sub-network conveniently learns to approximate sufficient summary statistics of the data from each block in the experiment~\citep{chen2021}.

In this work we develop a bespoke feed-forward neural network architecture, which allows us to efficiently deal with multiple blocks of data. The code we are providing is straightforwardly adaptable to other problems ``out of the box'', as the architecture can scale to multiple blocks of data and realistic complexity (dimensionality) of the observed data.
However, the overall approach is not restricted to this particular machine learning method; in fact, any sufficiently-flexible machine learning model may be used, as long as the loss function can be specified. In some settings, researchers may be interested in using machine learning models that only require little tuning, such as tree-based ensemble methods like Random Forests~\citep{breiman2001} or gradient boosted trees~\citep{friedman2001}.
On the other hand, when dealing with high-dimensional but structured data, such as in eye-tracking or neuroimaging ~\citep[or combinations of different types of data, see][]{turner2017approaches}, one may use other appropriate architectures, such as convolutional~\citep{lecun2015}, recurrent~\citep{rumelhart1986}, transformers, \citep{vaswani2017attention} or graph neural networks~\citep{zhou2020}. 
For further pointers to practical recommendations, see the code repository:~\url{https://github.com/simonvalentin/boed-tutorial}.

\subsection{Search over the space of possible experimental designs}
Once we have obtained an estimate of $U(\dbf)$ at a candidate design $\dbf$, we need to update the design appropriately when searching the design space to solve the experimental design problem $\dbf^\ast = \argmax_{\dbf} U(\dbf)$. Unfortunately, in the setting of discrete observed data $\ybf$ (such as choices between a set of options), gradients with respect to designs are generally unavailable, requiring the use of gradient-free optimization techniques. We here optimize the design $\dbf$ by means of Bayesian optimization~\citep[BO; see][for a review]{shahriari_taking_2015}, which has been used effectively in the experimental design context before~\citep[e.g.][]{martinez2014bayesopt, kleinegesse_bayesian_2020}. We specifically use a Gaussian Process (GP) as our probabilistic surrogate model with a Mat{\'e}rn-5/2 kernel and Expected Improvement as the acquisition function (these are standard choices).
A formal summary of our BOED approach is shown in Appendix 4.

For higher-dimensional continuous design spaces, this search may become difficult and necessitate the use of more scalable BO variants~\citep[e.g.][]{overstall2017, oh2018, Eduardo2023a}.
For discrete design spaces or combinations of discrete and continuous design spaces, one can straightforwardly parallelize the optimization for each level of the discrete variable. 
If the observed data are continuous, such as participants' response times, and one is able to compute gradients with respect to the designs (i.e., the simulator is differentiable), the design space may also be explored with gradient-based optimization~\citep{kleinegesse2021_gradbed}.
In some settings, we may have some non-trivial constraints on permissible experimental designs, and these can be enforced as part of the search, making sure that impermissible designs are rejected and/or not explored.
Thus, if we want our designs to have certain properties, we need to encode these constraints. 
For example, in our bandit tasks, we could enforce a certain minimum degree of randomness associated with each bandit arm's reward probability (though we do not do so in the present work), such that each reward probability is within in a specified range.

\begin{textbox}{Model discrimination: Simulation study} \label{box:5}
Our method reveals that the optimal reward probabilities for the model discrimination task are, approximately, $[0, 0, 0.6]$ for one block of trials and $[1, 1, 0]$ for the other block. These optimal designs stand in stark contrast to the usual reward probabilities used in such behavioral experiments, which characteristically use less extreme values~\citep{steyvers_bayesian_2009}. 
Contrary to our initial intuitions, and, presumably those of previous experiment designers, the extreme probabilities in these optimal designs are effective at distinguishing between particular theoretical commitments of our different models.
For instance, the AEG model (see Box~\ref{box:2}) can break ties between options with equivalent (observed) reward rates via the ``stickiness'' parameter. To illustrate this effect, consider the bandit with $[1, 1, 0]$ reward probabilities, as found above. 
Here, switching between the two options that always produce a reward is different from switching to the option that never produces a reward. 
Specifically, switching between the winning options can be seen as ``anti-stickiness'', motivated e.g., out of boredom or epistemic curiosity about subtle differences in the true reward rates of the winning options. This type of strategy is less apparent when the bandit arms are stochastic.
As such, these mechanisms can be investigated most effectively by not introducing additional variability due to stochastic rewards and instead isolating their effects.
Hence, we can interpret the, perhaps counter-intuitive, optimal experimental designs as effective choices for isolating distinctive mechanisms of our behavioral models or their parameters under the assumed priors. 

Using the neural network that was trained at that optimal design, we can cheaply compute approximate posterior distributions of the model indicator (as described in \emph{Materials and Methods}). This yields the confusion matrices in Figure~\ref{fig:md_post}, which show that our optimal designs result in considerably better model recovery than the baseline designs.
\begin{center}
    \includegraphics[width=0.5\linewidth]{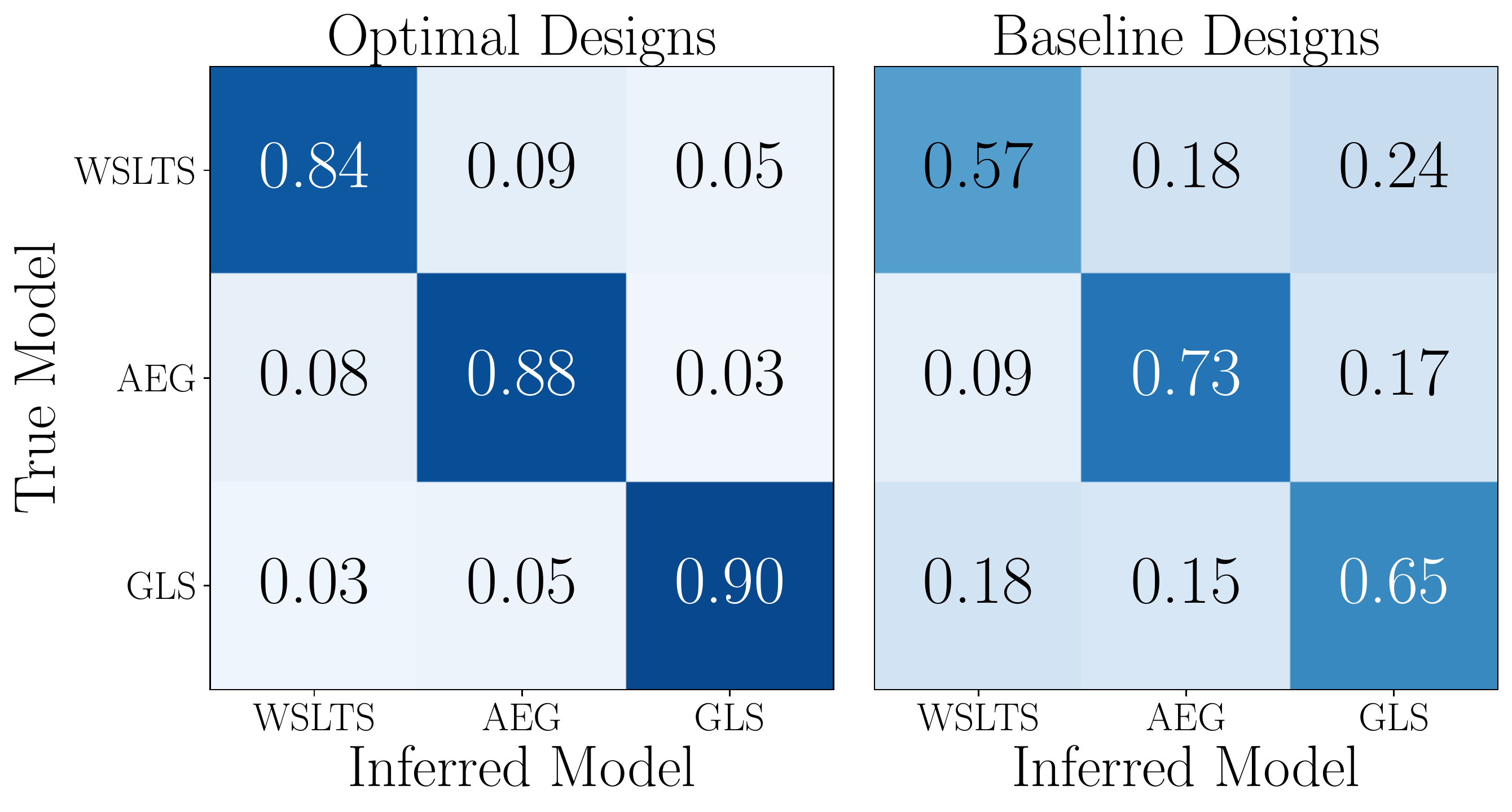}
    \captionof{figure}{Simulation study results for the model discrimination task, showing the confusion matrices of the inferred behavioral models, for optimal (left) and baseline (right) designs. Figure 4 has been adapted from Figure 2 in \citet{valentin2021bayesian}.} 
    \label{fig:md_post}
\end{center}
\end{textbox}

\section{Step 5: Validate the optimal experimental design in silico}
Once we have converged in our search over experimental designs, we can simulate the behavior of our theories under our optimal experimental design(s). Irrespective of the final experiment that will be run, model simulations can already provide useful information for theory building, e.g., by noticing that the simulator models generate implausible data for certain experimental designs.
This means that we can use simulations to assess whether we can recover our models, or individual model parameters, given the experimental design, as we did in our simulation study. 
We highlight this again as a critical step, since issues at this stage warrant revising the models under consideration and, importantly, any corresponding real-world experiment would be expected to yield uninformative data and waste resources~\citep{wilson2019ten}.

As we discuss in the \emph{Caveats} Section, BOED may sometimes lead to unusual yet effective experimental designs, and we suggest that surprising optimal designs may deserve careful consideration, as they may reveal subtle differences in predicted behaviors or model misspecifications. 
One appeal of optimal designs is that some of the ways of obtaining information about which models is correct correspond to finding situations in which one of them makes extreme predictions. 
BOED can thereby be seen as facilitating good prior predictive checking.
We have illustrated this step using our case study in Box~\ref{box:5} and Box~\ref{box:6}, where we present simulation study results for model discrimination and parameter estimation, respectively.

\begin{textbox}{Parameter estimation: Simulation study} \label{box:6}
We next discuss the parameter estimation results, focusing on the WSLTS model; the results for the AEG and GLS model can be found in Appendix 5. We find that the optimal reward probabilities for the WSLTS model are $[0, 1, 0]$, $[0, 1, 1]$ and $[1, 0, 1]$ for the three blocks. 
Similar to the model discrimination task, these optimal designs take extreme values, unlike commonly-used reward probabilities in the literature. 
In Figure~\ref{fig:pe_post} we show posterior distributions of the WSLTS model parameters for optimal and baseline designs.
We find that our optimal designs yield data that result in considerably improved parameter recovery, as compared to baseline designs.
\begin{center}
    \includegraphics[width=1\linewidth]{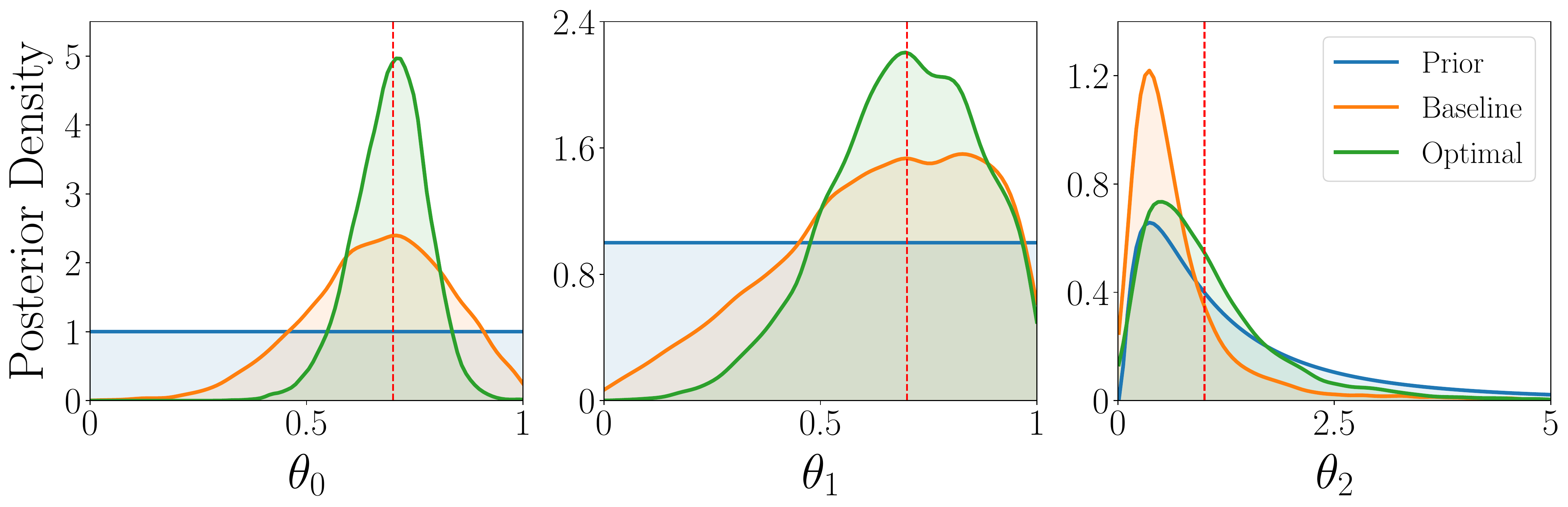}
    \captionof{figure}{Simulation study results for the parameter estimation task of the WSLTS model, showing the marginal posterior distributions of the three WSLTS model parameters for optimal (green) and baseline (orange) designs, averaged over $1{,}000$ simulated observations. The ground-truth parameter values were chosen in accordance with previous work on simpler versions of the WSLTS model~\citep{zhang_optimal_2010} and as plausible population values for the posterior reshaping parameter.}
    \label{fig:pe_post}
\end{center}
\end{textbox}

\section{Step 6: Run the real experiment}
Once we are satisfied with the found optimal experimental design, we can run the actual experiment. Here, the machine learning models we used to approximate MI allow us to cheaply perform posterior inference. After running the real-world experiment, we can use our trained ML models to cheaply compute approximate posterior distributions over models and their parameters.
This procedure provides a valuable addition to a Bayesian modeling workflow~\cite{schad2021toward, wilson2019ten}, as computing posterior distributions is a computationally expensive step, which is especially difficult for models with intractable likelihoods. We explain theoretical details below, and showcase an analysis of a real experiment with our case study in Boxes~\ref{box:7}$-$\ref{box:9}.

\subsection{Posterior Estimation} Once the neural network is trained, we can conveniently compute an approximate posterior distribution with a single forward-pass~\citep[see][for a derivation and further explanations]{kleinegesse_bayesian_2020}.
The NWJ lower bound is tight when $T_{\psib^{\ast}}(\bm{v}, \ybf) = 1 + \log p(\bm{v}|\ybf, \dbf^\ast) / p(\bm{v})$. By rearranging this, we can thus use our trained neural network $T_{\psib^{\ast}}(\bm{v}, \ybf)$ to compute a (normalized) estimate of the posterior distribution,
This can be done with the following equation,
\begin{equation} \label{eq:post}
    p(\bm{v}|\ybf, \dbf^\ast) = p(\bm{v}) e^{T_{\psib^{\ast}}(\bm{v}, \ybf) - 1}.
\end{equation}
Neural networks are notorious for large variations in performance, mainly due to the random initialization of network parameters. The quality of the posterior distribution estimated using Equation~\ref{eq:post} is therefore quite sensitive to the initialization of the neural network parameters. To overcome this limitation and to obtain a more robust estimate of the posterior distribution, we suggest to use \emph{ensemble learning} to compute Equation~\ref{eq:post}. This is done by training several neural networks ($50$ in our case study) and then averaging estimates of the posteriors obtained from each trained neural network.

\begin{textbox}{Human participant study: Methodology} \label{box:7}
To validate our method on empirical data, we collected a sample of $N=326$ participants from Prolific (\url{www.prolific.co}).~\footnote{All experiments were certified according to the University of Edinburgh Informatics Research Ethics Process, RT number 2019/58792.
}
Participants were randomly allocated to our optimal designs or to one of ten baseline designs (the same as in the simulation study); we refer to the former as the \emph{optimal} design group and to the latter as the \emph{baseline} group. The first two blocks correspond to the model discrimination task, i.e.~they were used to identify the model that matches a participant's behavior best. This was done by computing the posterior distribution over the model indicator $m$ using the trained neural networks (see Appendix 1) and then selecting the model with the highest posterior probability, i.e.~the \emph{maximum a posteriori} estimate.
This inference process was done online without a noticeable interruption to the experiment.
The following three blocks were then used for the parameter estimation task. Participants in the optimal design group were allocated the optimal design according to their inferred model (as provided by the model discrimination task).
Participants in the baseline group were again allocated to baseline designs. This data collection process facilitates gathering real-world data for both the model discrimination and parameter estimation tasks, allowing us to effectively compare our optimal designs to baseline designs.
See Appendix 2for more details about the human participant study setup.

Here, we follow the rationale that more useful experimental designs are those that lead to more information (or lower posterior uncertainty) about the variable of interest.
The belief about our variable of interest is provided by its posterior distribution, which we can estimate using the neural network output. We assess the quality of these posterior distributions by estimating their entropy, which quantifies the amount of \emph{information} they encode~\citep{shannon1948}. This is an effective and prominent metric to evaluate distributions, as it directly relates to the uncertainty about the variable of interest. 
For the model discrimination task, we specifically consider the Shannon entropy as a metric, since the model indicator $m$ is a discrete random variable.
As the continuous analogue of Shannon entropy, we use differential entropy for measuring the entropy of posterior distributions for the parameter estimation task, since the model parameters are continuous random variables. 
This allows us to quantitatively evaluate our optimal designs and compare their efficacy to that of baseline designs. 
\end{textbox}

\subsection{Summary statistics}
Additionally, as another component of our approach, we also obtain automatically-learned (approximate) sufficient summary statistics for data observed with the optimal experimental designs. Summary statistics are often crafted manually by domain experts, which tends to be a time-consuming and difficult task. In fact, hand-crafted summary statistics are becoming less effective and informative as our models naturally become more complex as science advances~\citep{chen2021}.
Our learned summary statistics could be used for various downstream tasks, e.g., as an input for approximate inference techniques (such as ABC) or as means to interpret and analyze the data more effectively, e.g., as a lower-dimensional representation used to visualize the data.

\begin{textbox}{Human participant study: Results} \label{box:8}
We here briefly present the results of our human participant study. For the model discrimination tasks, we find that our optimal designs yield posterior distributions that tend to have lower Shannon entropy than the baseline designs, as shown in Figure~\ref{fig:md_entropy}. Our ML-based approach thus allows us to obtain more informative beliefs about which model matches real individual human behavior best, as compared to baselines.

Having determined the best model to explain a given participant's behavior, we turn to the parameter estimation task, focusing our efforts on estimating the model parameters that explain their specific strategy. 
Similar to the model discrimination task, we here assess the quality of joint posterior distributions using entropy, where smaller values directly correspond to smaller uncertainty. 
Figure~\ref{fig:pe_entropy} shows the distribution of differential posterior entropies across participants that were assigned to the AEG simulator model; we show corresponding plots for the WSLTS and GLS models in Appendix 5. Similar to the model discrimination task, we find that our optimal designs provide more informative data, i.e.~they result in joint posterior distributions that tend to have lower differential entropy, as compared to the ones resulting from baseline designs, as shown in the figure.
\begin{center}
    \includegraphics[width=0.5\linewidth]{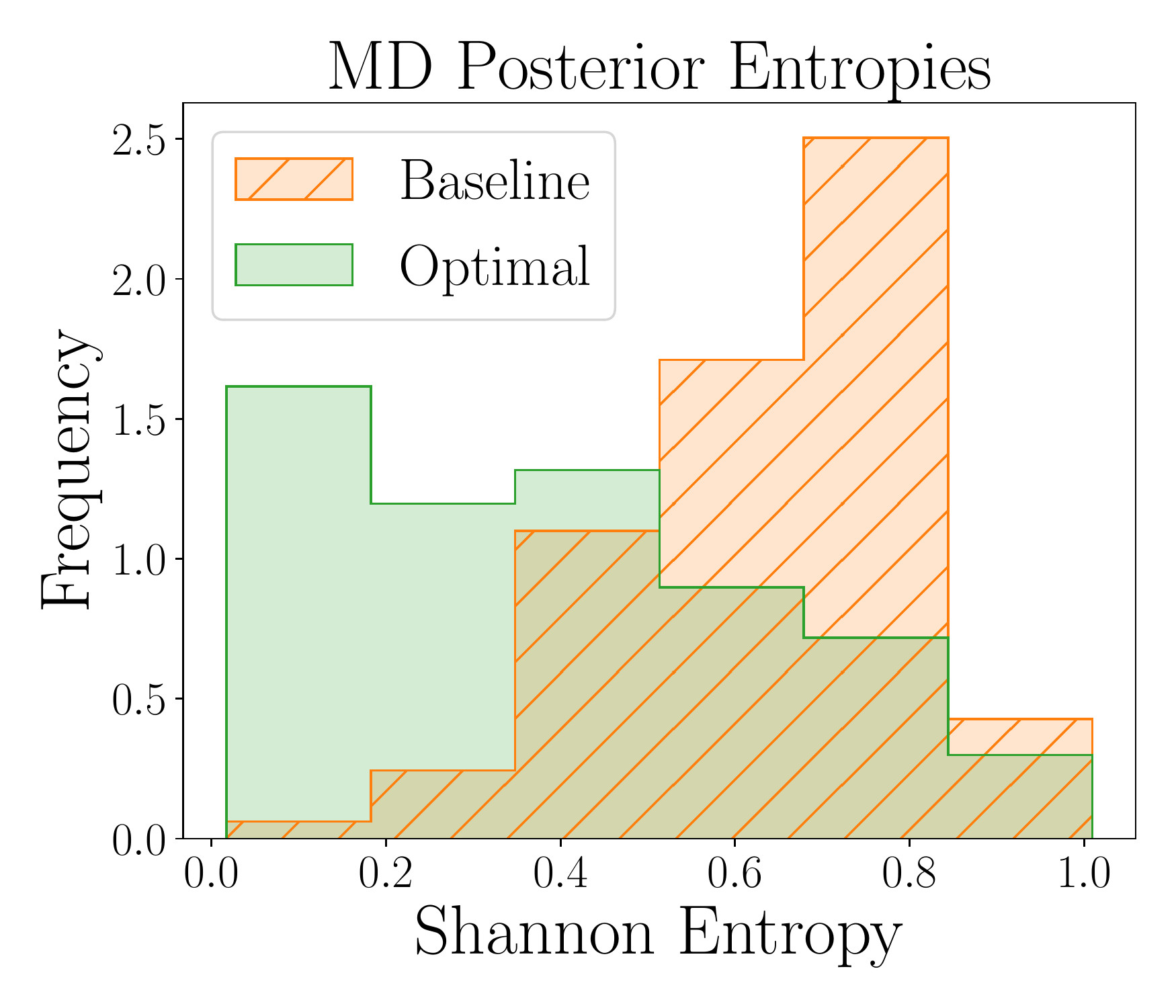}
    \captionof{figure}{Human-participant study results for the model discrimination (MD) task, showing the distribution of posterior Shannon entropies obtained for optimal (green) and baseline (orange) designs (lower is better). }
    \label{fig:md_entropy}
\end{center}
\begin{center}
    \includegraphics[width=0.5\linewidth]{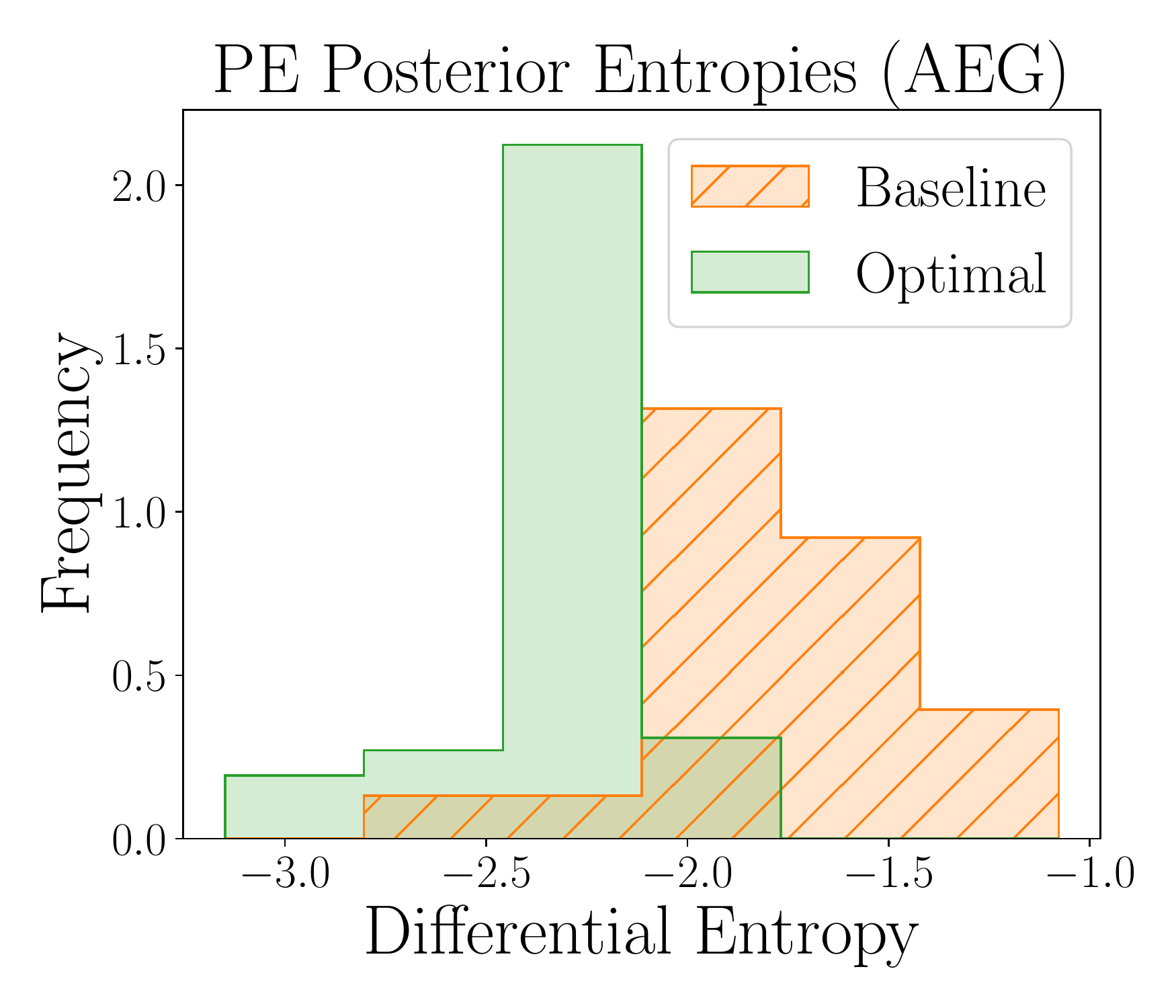}
    \captionof{figure}{Human-participant study results for the parameter estimation (PE) task of the AEG model, showing the distribution of posterior differential entropies obtained for optimal (green) and baseline (orange) designs (lower is better).}
    \label{fig:pe_entropy}
\end{center}

Moreover, the proportions of participants allocated to the WSLTS, AEG and GLS models for the optimal design were $62\,(37\%)$, $75\,(45\%)$ and $29\,(18\%)$, respectively. For the baseline designs, the proportions were $57\,(36\%)$, $22\,(14\%)$ and $81\,(51\%)$, for the WSLTS, AEG and GLS models respectively.
The finding that the largest proportion of participants for the optimal designs were best described by the AEG model differs from prior work, where most participants were best described by Win-Stay Lose-Shift (a special case of our WSLTS model)~\cite{steyvers_bayesian_2009}. 
This suggests that a reinterpretation of previous results is required, where the human tendency to stick with previous choices is less about myopically repeating the last successful choice, and more about balancing reward maximization, exploration, and a drive to be consistent over recent runs of choices.
Interestingly, we find different distributions over models for the optimal as compared to the baseline designs (A chi-square test revealed a significant difference between the distributions of WSLTS, AEG, and GLS in the Optimal and Baseline groups, $\chi^2(2) = 53.66$, $p < .0001$.).
In particular, while, descriptively, the proportions of participants allocated to the WSLTS model are very close ($37\%$, and $36\%$ for the optimal and baseline designs, respectively), the proportions allocated to the AEG and GLS model differ more substantially. 
This may suggest that people's strategy depends on the experimental design, potentially with some participants favoring strategies similar to the AEG model with more reliable reward environments, and the GLS in more unreliable (baseline) environments.
The models we have considered in our case study do not account for the possibility of transferring knowledge across experimental blocks. 
However, learning about the reward probabilities in one (or several) experimental blocks may influence people's expectations about the reward distribution of future block(s). 
For instance, having inferred (correctly or not), that at least one arm bandit arm always pays a reward, a learner would do well to quickly identify this arm in a future block~\citep[see, e.g.,][for a related line of research]{wang2018prefrontal}. 
Our space of computational models does not include such strategy switching mechanisms, and we discuss the impact of such potential misspecifications below in more detail.
While exploring these findings in detail from a cognitive science perspective is beyond the scope of the present work, we view this as a fruitful avenue for future work.

The extreme designs here may be surprising, considering the typically used stochastic rewards in bandit problems.
However, these designs allow the experiment to focus on particular aspects of people's behavior, which turns out to be most informative for distinguishing between the computational models we consider. 
For instance, while our AEG model performs uniformly random exploration, the WSLTS model performs uncertainty-guided exploration.
Meanwhile, our optimal designs raise the question of ecological validity, as it is unclear which kinds of ``reward environments'' are more reflective of real-world situations.
On the one hand, research in causal reasoning suggests that people expect relationships to be deterministic and reliable~\citep[e.g.,][]{schulz2006god}. 
On the other hand, some environments are known to be random and more akin to a classic, stochastic bandit problem, such as the stock market.
We view the question of the ecological validity of stochastic rewards as an important direction for future work, and discuss this more below in the \emph{Caveats} Section.
\end{textbox}

\begin{textbox}{Exploring model parameter disentanglement} \label{box:9}
In addition to measuring the efficacy of our optimal experiments via the differential entropy of posterior distributions, we can ask how well they isolate, or \emph{disentangle}, the effects of individual model parameters.
Specifically, we measure this by means of the (Pearson) correlation between model parameters in the joint posterior distribution, which we present in Figure~\ref{fig:avg_corrs} for all simulator models and for both optimal designs and baseline designs.
Here, a correlation of $r=0$ means that there is no (linear) relationship between the respective pair of inferred model parameters across individuals, thereby providing evidence that the data under this design allow us to separate the natural mechanisms corresponding to those parameters. On the other hand, correlations closer to $r=-1$ or $r=1$ would imply that the model parameters essentially encode the same mechanism. 
As shown descriptively in Figure~\ref{fig:avg_corrs}, our optimal designs are indeed able to isolate the mechanisms of individual model parameters more effectively than baseline designs.
Note that these are not confusion matrices, as displayed in Figure~\ref{fig:md_post}, but rather average correlation matrices of the model parameter posterior distributions. 
\begin{center}
    \includegraphics[width=0.8\linewidth]{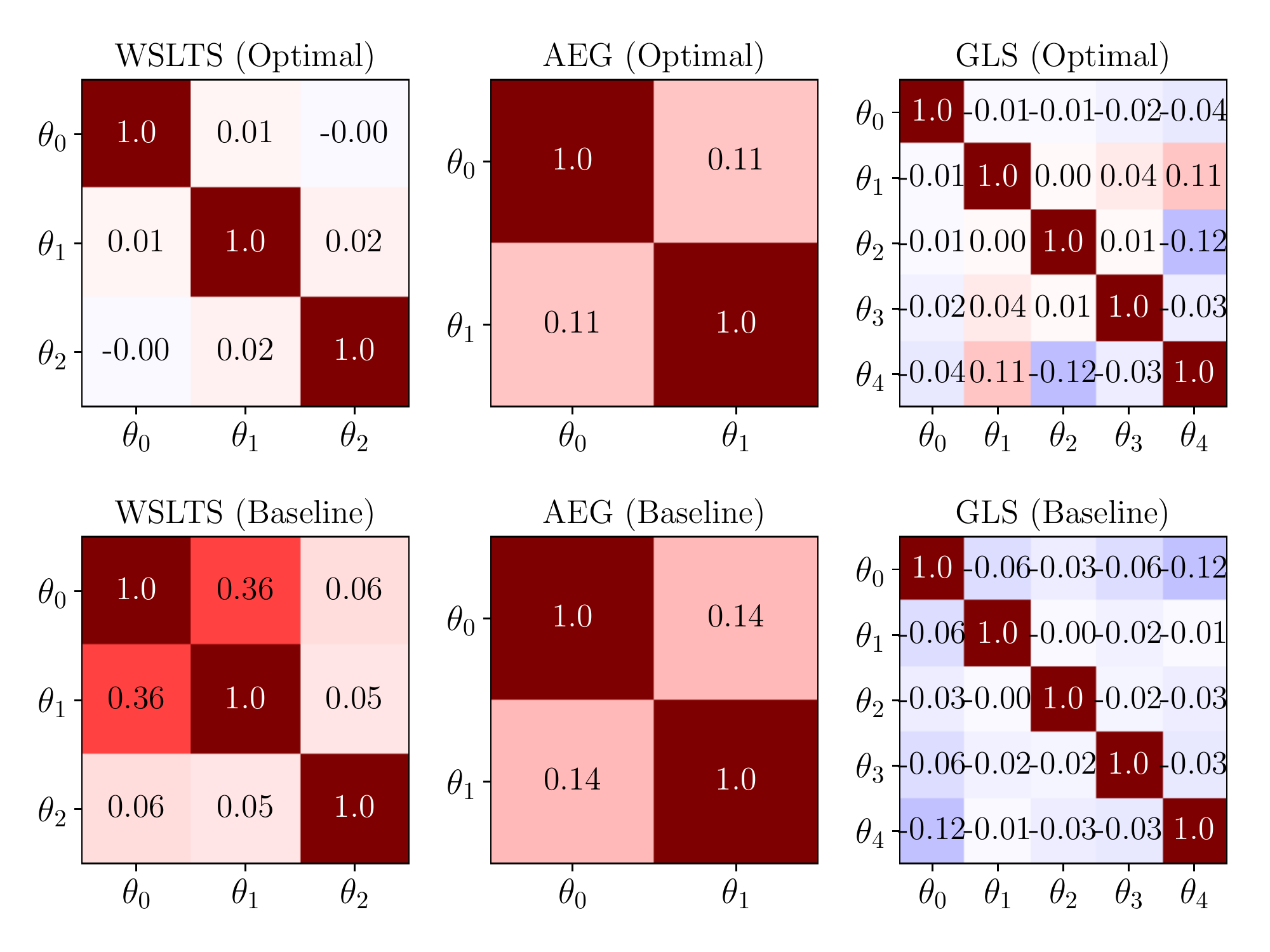}
    \captionof{figure}{Human-participant study average linear correlations in the posterior distribution of model parameters for all models in the parameter estimation task, and for both optimal (top) and baseline (bottom) designs.}
    \label{fig:avg_corrs}
\end{center}

To assess this observation quantitatively, we first $z$-transform~\citep{fisher1915frequency} the correlation coefficients (i.e., all entries below the main diagonal in the correlation matrices) and compute the average absolute $z$-score over the model parameters within each participant. 
The resulting average of absolute $z$-scores is an expression for the average magnitude of linear dependency between the posterior model parameters for each participant.
Next, we assess whether the average absolute $z$-scores are significantly different between those participants allocated to our optimal designs and those allocated to the baseline designs. We compute this statistical significance by conducting two-sided Welch's $t$-tests~\citep{welch1947generalization}.
We find that our optimal designs significantly outperform baseline designs for all three models, i.e.~for the WSLTS model we find a $p$-value of $p < 0.001$ ($t(60.12) = 13.89$), for the AEG model we have $p = 0.004$ ($t(27.30) = 3.12$) and for the GLS model we have $p < 0.001$ ($t(91.01) = 8.44$).
This provides additional evidence that our optimal designs are better able to disentangle the effects of individual model parameters than baseline designs. 
As explained earlier, we argue that this isolation of effects is facilitated by the extreme values of our optimal designs.
\end{textbox}

\subsection{Model checking}
As also emphasized by previous works on computational modeling, a model-based data analysis (such as via the amortized posterior distributions discussed above) should go hand-in-hand with model checking, also referred to as validation~\citep[e.g.,][]{wilson2019ten, palminteri2017importance}. 
A detailed treatment of methods and techniques for model checking is beyond the scope of this paper. 
However, it should be noted that model checking is just as important with data collected through optimized experimental designs as through hand-crafted designs.

\section{Caveats}
We now turn to more high-level considerations and discuss several categories of potential issues and caveats related to BOED.
We first discuss how the optimality of an experimental design is always defined with respect to a specific goal. Second, we consider how posterior inferences always depend on the experimental design and how this impacts BOED considerations. Third, we bring attention to the important issue of model misspecification in the context of BOED. Fourth, we discuss the topic of ecological validity and generalizability of our inferences with BOED. 
Finally, we cover the interpretability of our optimal designs. 
While we discuss these issues in light of BOED, they are just as important with hand-crafted experiments. Importantly, however, BOED makes these issues more salient, by turning the design of experiments into a more explicit process. 
Therefore, these points should be relevant to any researcher collecting or analyzing experimental data using computational models.

\subsection{No single design to rule them all}
The process of making all of our assumptions explicit and leaving the process of proposing concrete designs to the optimization procedure can provide us with assurance that our design is optimal with respect to our assumptions.
Therefore, this process may also reduce bias in the design of experiments, as the optimization process can easily be replicated. We believe that these properties will help tackle the notorious \emph{replication crisis}~\citep{ioannidis2005most,pashler2012, baker2016} and potentially setting new standards.

However, it is important to point out that the designs found using our optimization procedure are not assured to be optimal for all scenarios.
That is, there is no ``free lunch''~\cite{wolpert1996lack}: No design suits all situations, as the usefulness of experimental designs depends on the theories under consideration, prior beliefs and utility function. 
For instance, if we already have strong prior beliefs about a particular model parameter (such as people being strongly inclined to explore instead of exploit), there may be little to learn about this parameter and the BOED procedure may instead focus on parameters we have weaker prior beliefs about, which may yield different optimal designs.
More generally, different scientific goals can lead to different optimal designs. 

As an illustrative example, we consider the design of a survey item for political orientation. 
For simplicity, we assume a one-dimensional spectrum with extremes to either side of the scale being equally likely in the population, with most people falling in the middle of the scale.
In one scenario the goal may be to estimate a given person's political belief along the spectrum. 
Here, an informative question is likely going to be moderate, since extreme items on either side of the spectrum ignore the other side, have a low prior probability and are therefore less useful. 
However, if the goal is to distinguish ``extremist'' participants on one side of the spectrum from the rest of the population, or if the scientific question implies that a participant can only belong to either side of the one-dimensional political spectrum, then polarizing items are likely more informative than moderate items that do not reveal such extremist beliefs.
This emphasizes the importance of clearly defining the scientific goal, making sure that the computational models reflect the theories under consideration, as described in Steps 1 and 2. 

\subsection{Inferences are implicitly conditioned on the experimental design}
An important fact that is easy to forget is that all parameter inferences, be they Bayesian or frequentist, are inherently conditioned on the experimental designs used to generate the observed data. 
That is, the inferred belief about some parameter $\theta$ is determined by $p(\theta | \mathbf{y}, \mathbf{d})$ and not $p(\theta | \mathbf{y})$, where $\mathbf{d}$ is the design with which the data $\mathbf{y}$ was observed, even though this is usually not reflected in the notation explicitly, but rather kept implicit. 

As we have seen in our case study, different experimental designs may lead to different posterior inferences. 
For example, for the task of model discrimination, we found different proportions of participants best explained by our optimal designs and our baseline designs.
The question of when and why different experimental designs might support different models or theories is an important one that should concern anyone doing empirical research, but contrasting qualitatively different designs, as we do, makes this especially salient.
We would expect the simplest and most compelling evidence for some design $\mathbf{d}_a$ being ``better'' than another ($\mathbf{d}_b$) to be posterior distributions that are in qualitative agreement having conducted experiments with both designs, but with lower posterior variances given design $\mathbf{d}_a$. 
This raises the question of what any differences in inferences may imply. 
In extreme cases, one might observe two different experiments leading to dramatically different conclusions---whatever the provenance of the two experimental designs.
These differences may go so far as two experimental designs, $\mathbf{d}_a$ and $\mathbf{d}_b$ supporting different theories, such that $\mathbf{d}_a$ supports theory $A$, but falsifies theory $B$ and $\mathbf{d}_b$ supports theory $B$, but falsifies theory $A$. 
Such cases might plausibly be the result of serious model misspecification, where no model under consideration gives a good account of human behavior across different designs. 
For example, in the present case study, if every model under consideration supposes that people use fixed strategies, but in reality people adapt to the structure or statistics of the specific task at hand, then different designs would support different models. 
It seems plausible that some participants in bandit tasks might perform such strategy selection~\citep[e.g.,][]{lieder2017strategy}: Having engaged with a block of deterministic bandit arms, some participants may approach the second block differently than if they had just engaged with a more stochastic design, or may even switch strategies within the same block. 
The problem of model misspecification is thus brought to light very clearly by BOED, as can be seen in the case study. 

More generally, in cognitive science, we typically aspire to estimate parameters $\bm{\theta}$ that describe some stable (with respect to time, but also context) cognitive attribute or process. 
If two designs do not support the same estimated parameters, then $\bm{\theta}$ likely does not describe the stable trait that we would like it to describe. 
Sometimes, this may suggest adopting a higher-level model that decides when $\bm{\theta}$ should have an effect (such as a strategy switching model, as discussed above), such that our new $\bm{\theta}'$ describes a policy. 
Alternatively, we can explicitly limit the context (or domain) in which $\bm{\theta}$ should operate. 
For example, in the context of our case study, instead of generalizing our theory by proposing strategy-switching mechanisms, we may say that a theory on behavior in bandit problems should only cover stochastic reward environments, where no bandit arm has deterministic ($0$ or $1$) rewards. 
These considerations depend very much on the concrete phenomena being studied and the theories used to describe them. 

\subsection{Model misspecification \& informativeness}
Despite best efforts, computational models will suffer from some degree of model misspecification, failing to capture some aspects of people's behavior. 
While the misspecification may only be subtle in some cases, it can also be more severe, which poses a very real and important problem for computational modeling. 
BOED takes the perspective that the scientific question and models should come first and thus puts the model more into central focus than traditional approaches, thereby providing a more formal way to analyze the impact of modeling choices on the utility of different designs and expected results.
For instance, computational models may sometimes only provide reasonable predictions in limited regions of the space of possible experimental designs. 
We can inspect this in Step 5 of our proposed workflow and surprising predictions may motivate iterating on our theory/model building, or if we are committed to the model, can be exploited by BOED to perform powerful tests of competing theories.

While the issue of model misspecification is not specific to BOED, optimizing experimental designs based on misspecified models can have negative consequences. 
In particular, if the space of models is too narrow, our data may be less useful for later analyses with well-specified models. 
Two simple ways of lessening the impact of model misspecification are to enforce diversity constraints and to select qualitatively different high-utility designs to cover important behavioral phenomena~\citep{palminteri2017importance}. 
As discussed, we can include constraints in our search over experimental designs, such that our designs are not too narrowly focused on a single region of the parameter space. 
We could, for example, enforce this on a block-level, to ensure that every participant is presented with a diverse set of experimental designs.  
Alternatively, we can perform a more manual selection and choose designs that have high utility, but are likely to elicit qualitatively different phenomena. 
We present exemplary ways of engaging with the estimated utility function in Appendix 6.

\subsection{Ecological validity \& generalizability}
We have seen that BOED may sometimes lead to extreme or atypical designs, as the present results illustrate, and in some cases these designs may be deemed undesirable.
Similarly, we may sometimes want to use not only one experimental design for all units of observation, but, e.g., slightly vary the design across participants. 
As discussed by~\citet{yarkoni2022generalizability}, there is a common policy in experimental research of ignoring stimulus sampling variably, also referred to as the ``fixed-effect fallacy''~\cite{clark1973language}. This policy can lead to problems with generalizing inferences to broader classes of stimuli, especially when the phenomenon under question is likely sensitive to contextual effects~\citep{van2016contextual}. 
Fortunately, the BOED workflow is inherently designed to handle such cases. 

First, the optimization procedure over designs can include constraints (e.g., at most one deterministic bandit arm), which may be appropriate depending on the research question and can be viewed as inserting expert knowledge. 
Second, the approximation of the learned utility surface, allows for exploring the utility of different designs systematically. 
Rather than simply picking the design with maximal utility, practitioners can select from this surface to capture qualitatively different phenomena, or simply include slight variations of similar designs. 
For instance, researchers can pick the top-$k$ designs, or adopt a more sophisticated policy to ensure that the designs being used cover important parts of the design space to ensure robustness and generalizability of results, as argued for, e.g., by~\cite{yarkoni2022generalizability}.  
If the goal is to provide more varied designs, instead of the present approach of optimizing the reward probabilities directly, it would be straightforward to parametrize the design space differently and optimize a design policy. 
For instance, in the context of our case study, we could optimize the parameters ($\alpha$ and $\beta$) of a $\text{Beta}(\alpha, \beta)$ distribution from which the reward probabilities for the bandit arms of a given block are drawn. 
This is clearly only one option, and different ways of parametrizing design policies are possible, as an alternative or complement to choosing designs from the estimated utility surface. 
The BOED procedure is agnostic to such choices, and decisions should be based on the concrete problem being studied.
Researchers may here consider the trade-off between generalizabiltiy and variance control, which can be studied by simulating data from the model and performing parameter recovery analyses. We illustrate how the utility surface can be explored in Appendix 6, with code that shows how this can be accomplished in the related GitHub repository: \url{https://github.com/simonvalentin/boed-tutorial}. 

\subsection{Interpretability}
Simulating data from the computational models under consideration is of key importance in all stages of the experimental design process. 
In this work, we treat all models as simulators and, as argued, e.g., by~\citet{palminteri2017importance}, simulated data are crucial in assessing and potentially falsifying specific model commitments.
Prior predictive analyses are crucial for refining models and their priors, including detecting potential problems that may already be obvious from simulations alone, without having to run a real experiment. 
Beyond comparing models' posterior probability, posterior predictive analyses can provide insights into where models are ``right'' and where they may depart from human behavior, referring back to the previous theme of model misspecification. 
As proposed by~\citet{palminteri2017importance}, model comparison (through posterior probabilities) and falsification through simulation of behavioral effects thus serve complementary roles. 
Such simulations can serve as rigorous tests of theories, as a failure of a model to generate an effect of interest can be used as a criterion for rejecting the model (or a particular mechanism encoded in the model) in that form. 
These simulations should be considered a key part of the workflow presented in this work, beyond just inspecting posterior distributions (over models or their parameters). 

\section*{Conclusion and outlook}
This tutorial provides a flexible step-by-step workflow to finding optimal experimental designs, by leveraging recent advances in machine learning (ML) and Bayesian optimal experimental design (BOED). As a first step, the proposed workflow suggests to define a scientific goal, such as model discrimination or parameter estimation, thereby expressing the value of a particular design as a measurable quantity using a utility function. Secondly, the scientific theories under investigation need to be cast as a computational model from which we can sample synthetic data. The methodologies discussed in this tutorial make no assumption, however, on whether or not the likelihood function of the computational model needs be tractable, which opens up the space of scientific theories that can be tested considerably. In the third step, the design optimization problem needs to be set up, which includes deciding which experimental design variables need to be optimized and whether there are any additional constraints that should be included. The fourth step of our proposed workflow then involves using machine learning to estimate and optimize the utility function. In doing so, we discussed a recent approach that leverages neural networks to learn a mapping between experimental designs and expected information gain, i.e.~mutual information, and provided a novel extension that efficiently deals with behavioral data. In the fifth step, the discovered optimal designs are then validated \textit{in silico} using synthetic data. This ensures that the optimal designs found are sound and that no experimental resources are wasted, possibly also if it is found that the computational models need to be revisited after this step.
Lastly, in the sixth step, the real experiment can be performed using the discovered optimal design. Using the discussed methodologies, the trained neural networks can then conveniently be used to compute posterior distributions and summary statistics, immediately after the data collection.

This tutorial and the accompanying case study demonstrate the usefulness of modern ML and BOED methods for improving the way in which we design experiments and thus collect empirical data. 
Furthermore, the methodologies discussed in this tutorial optimize experimental designs for models of cognition without requiring computable likelihoods, or marginal likelihoods. 
This is critical to provide the methodological support required for scientific theories and models of human behavior that become increasingly realistic and complex.
As part of our case study, using simulations and real-world data, we showed that the proposed workflow and the discussed methodologies yield optimal designs that outperform human-crafted designs found in the literature, offering more informative data for model discrimination as well as parameter estimation.
We showed that adopting more powerful methodological tools allows us to study more realistic theories of human behavior, and our results provide empirical support for the expressiveness of the simulator models we studied in our case study.
BOED is currently an active area of research and future work will likely make some steps easier and more computationally efficient, such as automatic tuning of the machine learning methods used to estimate mutual information, variations of the loss function, improved ways of searching over the design space, new ways of dealing with model misspecification or computationally efficient sequential BOED~\citep{rainforth2023modern}. 
What will remain despite such technical advancements, however, is the need to formulate sound scientific questions and actively engaging with tools that help to automate designs. 
We thus believe that the steps and considerations outlined in the workflow presented in this work will stay relevant. 
More broadly, machine learning has seen success in behavioral research regarding the analysis of large datasets and discovering new theories with promising results~\cite[e.g.,][]{peterson2021using} but its potential for data collection and in particular experimental design has been explored remarkably little.
We view the present work as an important step in this direction.
%%%%%%%%%%%%%%%%%%%%%%%%%%%%%%%%%%%%%%%%%%%%%%%%%%%%%%%%%%%%

\section*{Acknowledgments} 
We thank Fiona Kn{\"a}ussel for her help in creating the schematics used in this paper, Maximilian Harkotte for valuable feedback during preparation of the paper and three anonymous reviewers for their careful reading of our manuscript and their many insightful comments and suggestions. SV was supported by a Principal's Career Development Scholarship, awarded by the University of Edinburgh. SK was supported in part by the EPSRC Centre for Doctoral Training in Data Science, funded by the UK Engineering and Physical Sciences Research Council (grant EP/L016427/1) and the University of Edinburgh.

\section*{Declaration}
The authors declare no competing interests.

% Bibliography
\bibliography{bib.bib}

\appendix
\begin{appendixbox}
\section*{Experiments}
\paragraph{Priors} We use a uniform categorical prior over the model indicator $m$, i.e.~$p(m) = \mathcal{U}(\{1, 2, 3\})$. We generally use uninformative priors $\mathcal{U}(0, 1)$ for all model parameters, except for the temperature parameter of the WSLTS model that has a $\text{LogNorm}(0, 1)$ prior, as it acts as an exponent in reshaping the posterior. We generate $50{,}000$ samples from the prior and then simulate corresponding synthetic data $\ybf | \thetab, \dbf$ at every design $\dbf$.

\paragraph{Sub-networks} For all of our experiments, we use sub-networks $\mathbf{S}_{\psib}(\bm{v}, \ybf)$ that consist of two hidden layers with 64 and 32 hidden units, respectively, and ReLU activation functions. The number of sufficient statistics we wish to learn for each block of behavioral data is given by the number of dimensions in the output layer of the sub-networks. These are $6$, $8$, $6$ and $8$ units for the model discrimination, parameter estimation (WSLTS), parameter estimation (AEG) and parameter estimation (GLS) experiments, respectively. The flexibility of a sub-network is naturally improved when increasing the number of desired summary statistics, but the computational cost increases accordingly. When the number of summary statistics is too low, the summary statistics we learn may not be sufficient. We have found the above numbers of summary statistics to be effective middle-grounds and refer to~\cite{chen2021} for more detailed guidance on how to select the number of summary statistics.

\paragraph{Main Network} The main network $T_{\psib}(\bm{v}, \ybf)$ consists of the concatenated outputs of the sub-networks for each block of behavioral data and the variable of interest. This is then followed by two fully-connected layers with ReLU activation functions. For the model discrimination experiment we use $32$ hidden units for the two hidden layers, while we use $64$ and $32$ hidden units for the parameter estimation experiments. See Figure~\ref{fig:nn_arch} for a visualization of this bespoke neural network architecture.

\paragraph{Training} We use the Adam optimizer to maximize the lower bound shown in Equation~\ref{eq:lb}, with a learning rate of $10^{-3}$ and a weight decay of $10^{-3}$ (except for the parameter estimation (WSLTS) experiments where we use a weight decay of $10^{-4}$). We additionally use a plateau learning rate scheduler with a decay factor of $0.5$ and a patience of $25$ epochs. We train the neural network for $200$, $400$, $300$ and $300$ epochs for the model discrimination, parameter estimation (WSLTS), parameter estimation (AEG) and parameter estimation (GLS) experiments, respectively. At every design we simulate $50{,}000$ samples from the data-generating distribution (one for every prior sample) and randomly hold out $10{,}000$ of those as a validation set, which are then used to compute an estimate of the mutual information via Equation~\ref{eq:lb}. During the BO procedure we select an experimental budget of $400$ $U(\dbf)$ evaluations ($80$ of which were initial evaluations), which is more than double needed to converge.

\end{appendixbox}

\begin{appendixbox}
\section*{Human Participant Study}
\paragraph{Task}
Participants completed the multi-armed bandit tasks in online experiments. 
After going through instructions on the interface and setup of the task, participants were required to pass a series of 5 comprehension questions (in a true-false format).
If any of the comprehension questions were answered incorrectly, the participant was sent back to the instructions and could only progress to the task once they answered all questions correctly. 
An example screen-shot of the task interface can be found in Appendix 5.

\paragraph{Two-Phase Design}
To allocate participants to the parameter estimation designs for the respective model that best matched their behavior in the model discrimination blocks, we implemented a simple API that uses the ensemble of trained neural networks and performs approximate MAP inference over the model indicator (see the previous sub-section on posterior estimation). 
As we have obtained amortized posterior distributions as by-products in the BOED procedure, this inference can be done efficiently: Forward-passes through the neural networks are computationally cheap, only taking a fraction of a second, which means that there was no noticeable delay for the allocation of the optimal model for each participant.

\paragraph{Participants}
$N=326$ adults ($154$ female, mean age $M=35.61$ $SD = 12.59$) participated in the experiment in return for a basic payment of \pounds$0.60$ and performance related bonuses of up to \pounds$1.00$. Participants took, on average, $8.8$ ($SD = 2.89$) minutes to complete the task.
\end{appendixbox}

\begin{appendixbox}
\section{Computational models}
We here provide further details about the computational models of human behavior in bandit tasks, which we briefly presented in the main text. We consider the multi-armed bandit setting where, at each trial $t=1, \dots, T$, participants have to make an action $a^{(t)} \in \{1, \dots, K\}$, which consists of choosing any of the $K$ bandit arms, and subsequently observe a reward $r^{(t)} \in \{0, 1\}$. After a participant has gone through all $T$ trials, we summarize their behavior by a set of actions $\bm{a} = (a^{(1)}, \dots, a^{(T)})$ and observed rewards $\bm{r} = (r^{(1)}, \dots, r^{(T)})$.

Whether a participant observes a reward of $0$ or $1$ when making a particular choice $a^{(t)}$ depends on the specified reward probability.
We here assume that each bandit arm $k \in \{1, \ldots, K\}$ is associated with a Bernoulli reward distribution. Each of these reward distributions has a (potentially) different reward probability, which is given by the corresponding entry in the design vector $\bm{d} \in [0, 1]^K$. In other words, the reward $r^{(t)}$ a participant receives when making a particular choice $a^{(t)} = k$ is sampled via $r^{(t)} \sim \text{Bernoulli}(d_k)$, where $d_k$ is the $k$-th element of $\dbf$.

Computational models of human behavior in bandit tasks only differ in how they model the choices of a participant depending on the previous actions and rewards. To help us describe the mechanisms of such computational models, we here define the vectors $\bm{\alpha}$ and $\bm{\beta}$, which store the number of observed $0$ and $1$ rewards for all $K$ arms. 
That is, $\alpha_k$ refers to the number of times the $k$-th arm was selected and subsequently generated a reward. Similarly, $\bm{\beta}_k$ refers to the number of times a participant did not observe a reward when selecting the $k$-th bandit arm. Below, we describe each of the computational models in more detail and include corresponding pseudo-code.

\subsection{Win-Stay Lose-Thompson-Sample (WSLTS)}
Here, we propose Win-Stay Lose-Thompson-Sample (WSLTS) as an amalgamation of Win-Stay Lose-Shift~\citep[WSLS;][]{robbins_aspects_1952} and Thompson Sampling~\citep{thompson1933likelihood}. 
WSLTS has three model parameters $(\theta_0, \theta_1, \theta_2)$ that regulate different aspects of exploration and exploitation behavior. Specifically, $\theta_0$ corresponds to the probability of staying after winning, i.e.~the probability of re-selecting the previous arm $a^{(t-1)}$ after having observed $r^{(t-1)} = 1$. However, with probability $1 - \theta_0$ the agent may decide to choose a different bandit arm even after having observed a reward of $1$ at the previous trial. In this case, the agent performs Thompson Sampling, using a temperature parameter $\theta_2$, to decide which bandit arm to select (see Equation~\ref{eq:wslts}). Conversely, $\theta_1$ corresponds to the probability of switching to another arm when observing a loss in the previous trial, i.e.~when $r^{(t-1)} = 0$, in which case the agent again performs Thompson Sampling to select another bandit arm. However, the agent may also re-select the previous bandit arm with probability $1 - \theta_1$ even after having observed a loss.

During the exploration phases mentioned above, the WSLTS agent performs Thompson Sampling from a reshaped posterior, which is controlled via a temperature parameter $\bm{\theta}_2$. The corresponding choice of bandit arm is then given by
\begin{equation}
a^{(t)} = \argmax_{k} \, \omega_k, \quad \omega_{k} \sim 
\begin{cases}
\text{Beta}\left(\alpha_{k}^{1/\bm{\theta}_2},  \beta_{k}^{1/\bm{\theta}_2}\right) &\text{if $k \neq a^{(t-1)}$ }\\
0 &\text{if $k = a^{(t-1)}$ },
\end{cases} 
\label{eq:wslts}
\end{equation}
where $\alpha_k$ and $\beta_k$ correspond to the number of times a participant observed a reward of $1$ and $0$, respectively, for a bandit arm $k$. This stands in contrast to standard WSLS, where the agent shifts to another arm uniformly at random.
We note that, for $\theta_2$ values close to $1$ we recover standard Thompson-Sampling (excluding the previous arm), while for $\theta_2 \rightarrow \infty$ we recover standard WSLS, and for $\theta_2 \rightarrow 0$ we obtain a greedy policy. We provide pseudo-code for the WSLTS computational model in Algorithm~\ref{algo:WSLTS}.

\begin{center}
\begin{minipage}{.64\linewidth}
\begin{algorithm}[H]
\caption{Win-Stay Lose-Thompson-Sample (WSLTS)}\label{algo:WSLTS}
\begin{algorithmic}[1]
\Input Parameter $\bm{\theta}=(\theta_0, \theta_1, \theta_2)$, design $\bm{d}=(d_1, \dots, d_K)$
\Output Actions $\bm{a}=(a^{(1)}, \dots, a^{(T)})$, rewards $\bm{r} = (r^{(1)}, \ldots, r^{(T)})$
\State Initialize pseudo-counts $\alpha_k$ and $\beta_k$ to $1$ for all bandit arms $k$. 
\For{t = 1, \ldots, T}
    \State {Sample $u \sim \mathcal{U}(0, 1)$.}
    \If{t = 1}
        \State {Select the first bandit arm $a^{(1)}$ uniformly at random.}
    \ElsIf{$r^{(t-1)} = 1$}
        \If{$u < \theta_0$}
            \State {Re-select the previous bandit arm $a^{(t-1)}$.}
        \Else
            \State{Thomson-Sample according to equation \ref{eq:wslts}}.
        \EndIf
    \Else
        \If{$u < \theta_1$}
            \State{Thomson-Sample according to equation \ref{eq:wslts}.}
        \Else
            \State {Re-select the previous previous bandit arm $a^{(t-1)}$.}
        \EndIf
    \EndIf
    \State {Sample the reward $r^{(t)} \sim \text{Bernoulli}(d_{a^{(t)}})$.}
    \State{Increment $\alpha_{a^{(t)}} \leftarrow \alpha_{a^{(t-1)}} +1$ if $r^{(t)} = 1$ and $\beta_{a_t} \leftarrow \beta_{a^{(t-1)}} +1$ if $r^{(t)} = 0$.}
\EndFor
\end{algorithmic}
\end{algorithm}
\end{minipage}
\end{center}

\subsection{Autoregressive $\varepsilon$-Greedy (AEG)}
Standard $\varepsilon$-Greedy~\citep[e.g.,][]{sutton_reinforcement_2018} is a ubiquitous method in reinforcement learning, where the agent selects the arm with the highest expected reward with probability $1-\varepsilon$, where $\varepsilon$ is a model parameter. Conversely, with probability $\varepsilon$, the agent performs exploration by uniformly selecting a bandit arm.
Here, we propose Auto-regressive $\varepsilon$-Greedy (AEG) as a generalization of
$\varepsilon$-Greedy, which allows for modeling people's tendency towards auto-regressive behavior~\citep[e.g.,][]{gershman2020origin}.

Specifically, the AEG model has two model parameters $(\theta_0, \theta_1)$, where $\theta_0$ corresponds to the same $\varepsilon$ parameter as in $\varepsilon$-Greedy. However, as opposed to randomly selecting any bandit arm, the probability of selecting the previous arm, in order to break ties between options with the same expected reward, is specifically controlled via the second model parameter $\theta_1$. We provide pseudo-code for the AEG computational model in Algorithm~\ref{algo:AEG}.
\begin{center}
\begin{minipage}{.66\linewidth}
\begin{algorithm}[H]
\caption{Autoregressive $\varepsilon$-Greedy (AEG)}\label{algo:AEG}
\begin{algorithmic}[1]
\Input Parameter $\bm{\theta}=(\theta_0, \theta_1)$, design $\bm{d}=(d_1, \dots, d_K)$
\Output Actions $\bm{a}=(a^{(1)}, \dots, a^{(T)})$, rewards $\bm{r} = (r^{(1)}, \ldots, r^{(T)})$
\State Initialize pseudo-counts $\alpha_k$ and $\beta_k$ to $1$ for all bandit arms $k$. 
\For{t = 1, \ldots, T} 
    \State {Sample $u \sim \mathcal{U}(0, 1)$ and $v \sim \mathcal{U}(0, 1)$.}
    \State {Update the estimated reward probability for all $K$ bandit arms as $\frac{\alpha_k}{\alpha_k + \beta_k}$.}
    \State {Let $M$ be the set of bandit arms with maximal expected reward.}
    \If{t = 1}
        \State {Select the first bandit arm $a^{(1)}$ uniformly at random.}
    \ElsIf{$u < \theta_0$}
        \If{$v < \bm{\theta}_1 + \left(1 - \bm{\theta}_1\right) / K$}
            \State {Re-select the previous bandit arm $a^{(t-1)}$.}
        \Else
            \State {Randomly select another bandit arm.}
        \EndIf
    \Else 
        \If{$a^{(t-1)} \in M$ \textbf{and} $v < \bm{\theta}_1 + \left(1 - \bm{\theta}_1\right) / |M|$}
            \State {Re-select the previous bandit arm $a^{(t-1)}$.}
        \Else
            \State {Randomly select another bandit arm among the set $M$.}
        \EndIf
    \EndIf
    \State {Sample the reward $r^{(t)} \sim \text{Bernoulli}(d_{a^{(t)}})$.}
    \State{Increment $\alpha_{a^{(t)}} \leftarrow \alpha_{a^{(t-1)}} +1$ if $r^{(t)} = 1$ and $\beta_{a_t} \leftarrow \beta_{a^{(t-1)}} +1$ if $r^{(t)} = 0$.}
\EndFor
\end{algorithmic}
\end{algorithm}
\end{minipage}
\end{center}

\subsection{Generalized Latent State (GLS)}
\citep{lee_psychological_2011} proposed a latent state model for bandit tasks whereby a learner can be in either an \emph{explore} or an \emph{exploit} state and switch between these as they go through the task. 
Here, we propose the Generalized Latent State (GLS) model, which unifies and extends latent-state and latent-switching models,  previously studied in \citep{lee_psychological_2011}, allowing for more flexible and structured transitions.~\footnote{We thank Patrick Laverty for contributing towards this.}
The transition distribution over whether the agent is in an exploit state is specified by four parameters, each of which specify the probability of being in an exploit state at trial $t$. This transition distribution is denoted by $\pi(l, r_{t-1})$, where $l$ is the previous latent state and $r_{t-1}$ the reward observed in the previous trial.
We here refer to trials where a bandit arm was selected but failed to produce a reward as \emph{failures}. We provide pseudo-code for our GLS computational model in Algorithm~\ref{algo:GLS}.

\begin{center}
\begin{minipage}{.75\linewidth}
\begin{algorithm}[H]
\caption{Generalized Latent State (GLS)}\label{algo:GLS}
\begin{algorithmic}[1]
\Input Parameter $\bm{\theta}=(\theta_0, \theta_1, \theta_2, \theta_3, \theta_4)$, design $\bm{d}=(d_1, \dots, d_K)$
\Output Actions $\bm{a}=(a^{(1)}, \dots, a^{(T)})$, rewards $\bm{r} = (r^{(1)}, \ldots, r^{(T)})$
\State Initialize pseudo-counts $\alpha_k$ and $\beta_k$ to $1$ for all bandit arms $k$.
\State Sample the initial latent state $l \sim \text{Bernoulli}(0.5)$.
 
\For{t = 1, \ldots}
    \State {Let $R \leftarrow$ all $k$ such that $k = \argmax_{k} \bm{\alpha}$ (arms with the maximal number of rewards).}
    \State {Let $F \leftarrow$ all $k$ such that $k = \argmin_{k} \bm{\beta}$ (arms with the minimal number of failures).}
    \State {Let $S \leftarrow R \cap F$.}
    \State {Sample $u \sim \mathcal{U}(0, 1)$ and $v \sim \mathcal{U}(0, 1)$.}
    \If{t = 1}
        \State {Select the first bandit arm $a^{(1)}$ uniformly at random.}
    {\ElsIf{$|S| > 1$} 
        \State {Randomly select a bandit arm $a^{(t)}$ from the set $S$.}
    \ElsIf{$|S| = 1$}
        \If{$u < \theta_0$}
            \State {Select the bandit arm $a^{(t)} \in S$}
        \Else
            \State {Randomly select another bandit arm.}
        \EndIf
    \ElsIf{ $v < \pi(l, r_{t-1})$} 
     \State {$l \leftarrow 1$}\Comment{Latent state is \emph{exploit}}
       \If{$u < \theta_0$}
            \State {Let $R_{\text{min}}$ be the set of bandit arms in $R$ with the minimal number of failures.}
            \State {Randomly select a bandit arm $a^{(t)}$ from the set $R_{\text{min}}$.}
        \Else
            \State {Randomly select another bandit arm.}
        \EndIf
    \Else
     \State {$l \leftarrow 0$} \Comment{Latent state is \emph{explore}}
        \If{$u < \theta_0$}
            \State {Let $F_{\text{max}}$ be the set of bandit arms in $F$ with the maximal number of rewards.}
            \State {Randomly select a bandit arm $a^{(t)}$ from the set $F_{\text{max}}$.}
        \Else
            \State {Randomly select another bandit arm.}
        \EndIf
    \EndIf}
    \State {Sample the reward $r^{(t)} \sim \text{Bernoulli}(d_{a^{(t)}})$.}
    \State{Increment $\alpha_{a^{(t)}} \leftarrow \alpha_{a^{(t-1)}} +1$ if $r^{(t)} = 1$ and $\beta_{a_t} \leftarrow \beta_{a^{(t-1)}} +1$ if $r^{(t)} = 0$.}
\EndFor
\end{algorithmic}
\end{algorithm}
\end{minipage}
\end{center}
\end{appendixbox}

\begin{appendixbox}
\section{BOED Algorithm}

Algorithm~\ref{algo:minebed} describes the BOED procedure used in our work to determine optimal experiments in bandit tasks.

\begin{center}
\begin{minipage}{0.65\linewidth}
\begin{algorithm}[H]
\caption{BOED}\label{algo:minebed}
\begin{algorithmic}[1]
\Input Simulator model $\ybf \sim p(\ybf|\bm{v}, \dbf)$, prior distribution $p(\bm{v})$, neural network (NN) architecture for $T_{\psib}(\bm{v}, \ybf)$
\Output Optimal design $\dbf^\ast$, trained NN $T_{\psib^{\ast}}(\bm{v}, \ybf)$ at $\dbf^\ast$
\State Randomly initialize the experimental designs $\dbf_n \leftarrow \dbf_0$
\State Initialize the Gaussian Process for Bayesian optimization (BO)
\While {$U(\dbf_n)$ not converged}
    \State {Sample from the prior: $\bm{v}^{(i)} \sim p(\bm{v})$ for $i=1, \dots, M$}
    \State {Sample from the simulator: $\ybf^{(i)} \sim p(\ybf|\bm{v}^{(i)}, \dbf)$ for $i=1, \dots, M$}
    \State {Randomly initialize the NN parameters $\psib_n \leftarrow \psib_0$}
    \While {$U(\dbf_n; \psib_n)$ with fixed $\dbf_n$ not converged}
        \State {Compute a sample average of the lower bound (see the main text)}
        \State {Estimate gradients of the sample average with respect to $\psib_n$}
        \State {Update $\psib_{n}$ using any gradient-based optimizer}
    \EndWhile
    \State {Use $\dbf_n$ and $U(\dbf_n)$ to update the Gaussian Process}
    \State {Use BO to determine which $\dbf_{n+1}$ to evaluate next}
\EndWhile
\end{algorithmic}
\end{algorithm}
\end{minipage}
\end{center}
\end{appendixbox}

\begin{appendixbox}
\section{Additional figures and results}

In this section, we provide additional figures and results that supplement our main text. Figure~\ref{fig:exp_task} shows a screenshot of the bandit task that participants were presented with in our human experiment. Figures~\ref{supfig:pe_sim_posts_aeg} and~\ref{supfig:pe_sim_posts_gls} show the parameter estimation results for the AEG and GLS simulator models, respectively, of our simulation study. Specifically, these figures show marginal posterior distributions of the model parameters, averaged over several (simulated) observations, for baseline designs and our optimal designs. With regards to the real experiment with human participants, Figures~\ref{fig:pe_real_entropy_wslts} and~\ref{fig:pe_real_entropy_gls} show distributions of the posterior entropies in the parameter estimation task for the WSLTS and GLS simulator models, respectively. These distributions compare the differential entropy of the posteriors obtained using our optimal designs with those obtained using the baseline designs. Lastly, Figures~\ref{fig:pe_real_post_wslts_ex1} and~\ref{fig:pe_real_post_wslts_ex2} show example posteriors of human participants assigned to the WSLTS model, while Figures~\ref{fig:pe_real_post_gls_ex1} and~\ref{fig:pe_real_post_gls_ex2} show example posteriors of human participants assigned to the GLS model.

\begin{center}
    \includegraphics[width=0.8\linewidth]{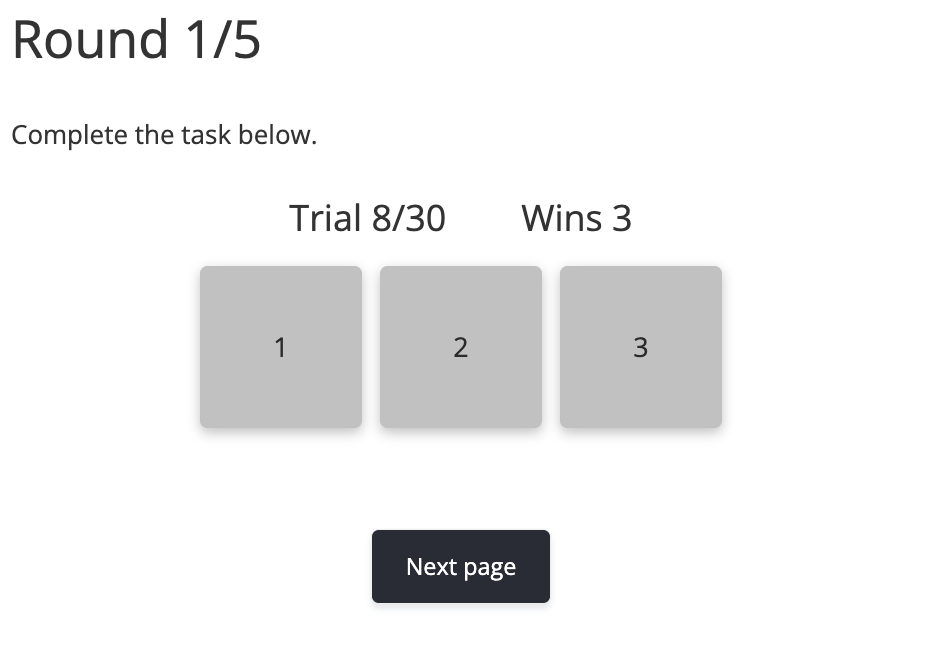}
    \captionof{figure}{Screenshot of bandit task that participants completed in online experiments.}
    \label{fig:exp_task}
\end{center}

\begin{center}
    \includegraphics[width=0.7\linewidth]{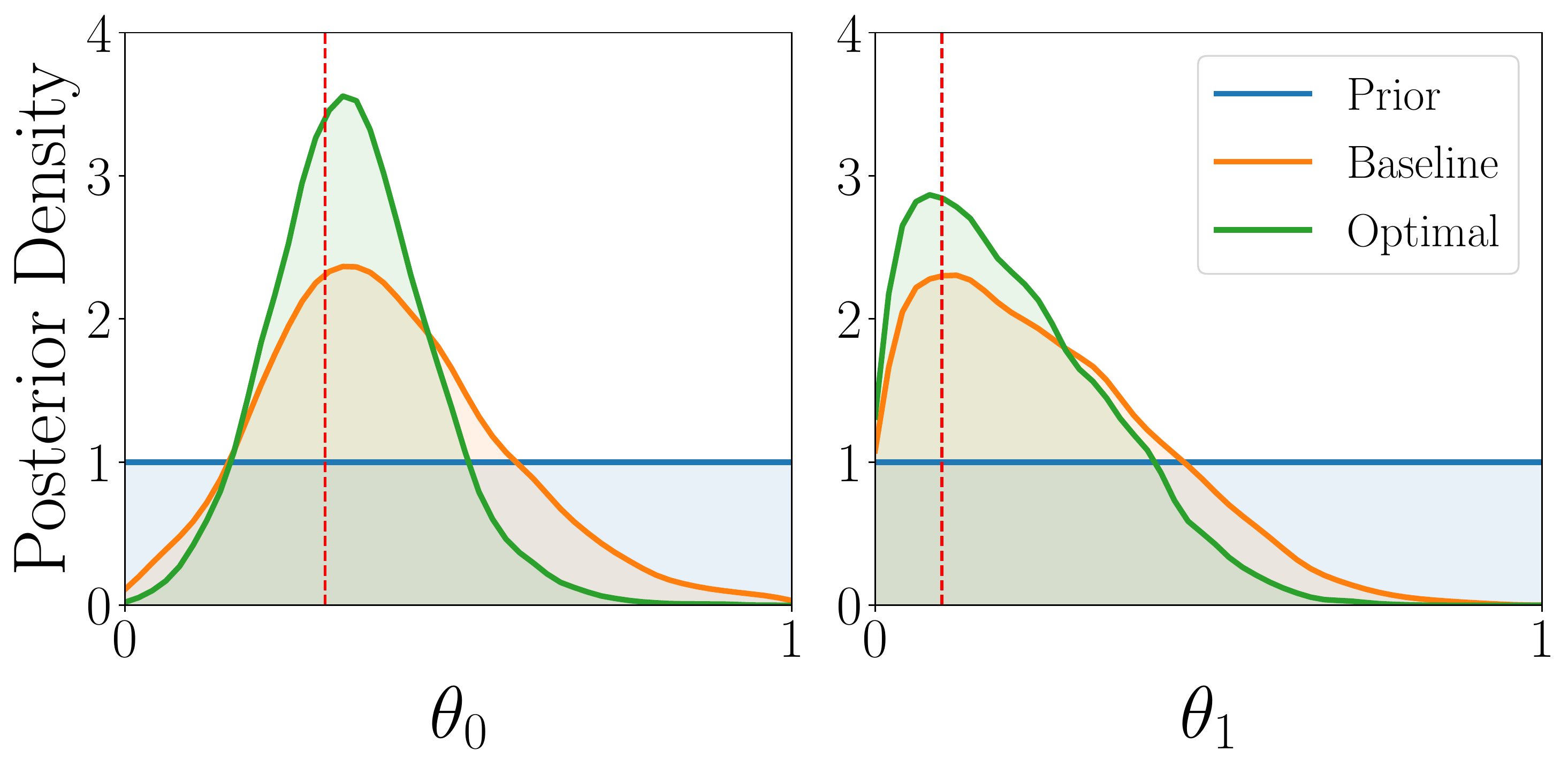}
    \captionof{figure}{Results for the parameter estimation task of the AEG model, showing the marginal posterior distributions of the three AEG model parameters for optimal (green) and baseline (orange) designs, averaged over $1{,}000$ observations.}
    \label{supfig:pe_sim_posts_aeg}
\end{center}

\begin{center}
    \includegraphics[width=\linewidth]{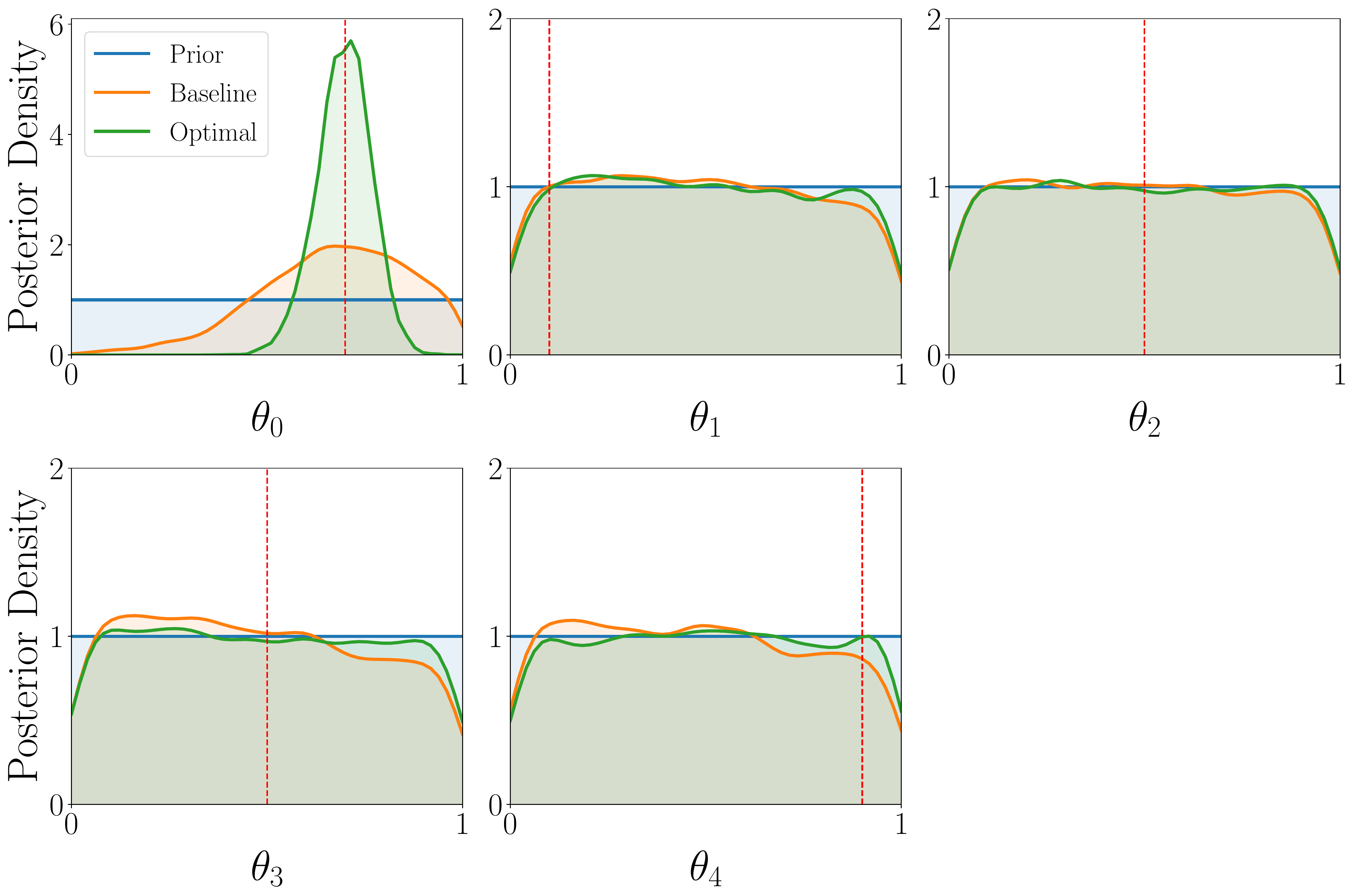}
    \captionof{figure}{Results for the parameter estimation task of the GLS model, showing the marginal posterior distributions of the three GLS model parameters for optimal (green) and baseline (orange) designs, averaged over $1{,}000$ observations.}
    \label{supfig:pe_sim_posts_gls}
\end{center}

\begin{center}
\begin{minipage}{0.48\linewidth}
    \includegraphics[width=\linewidth]{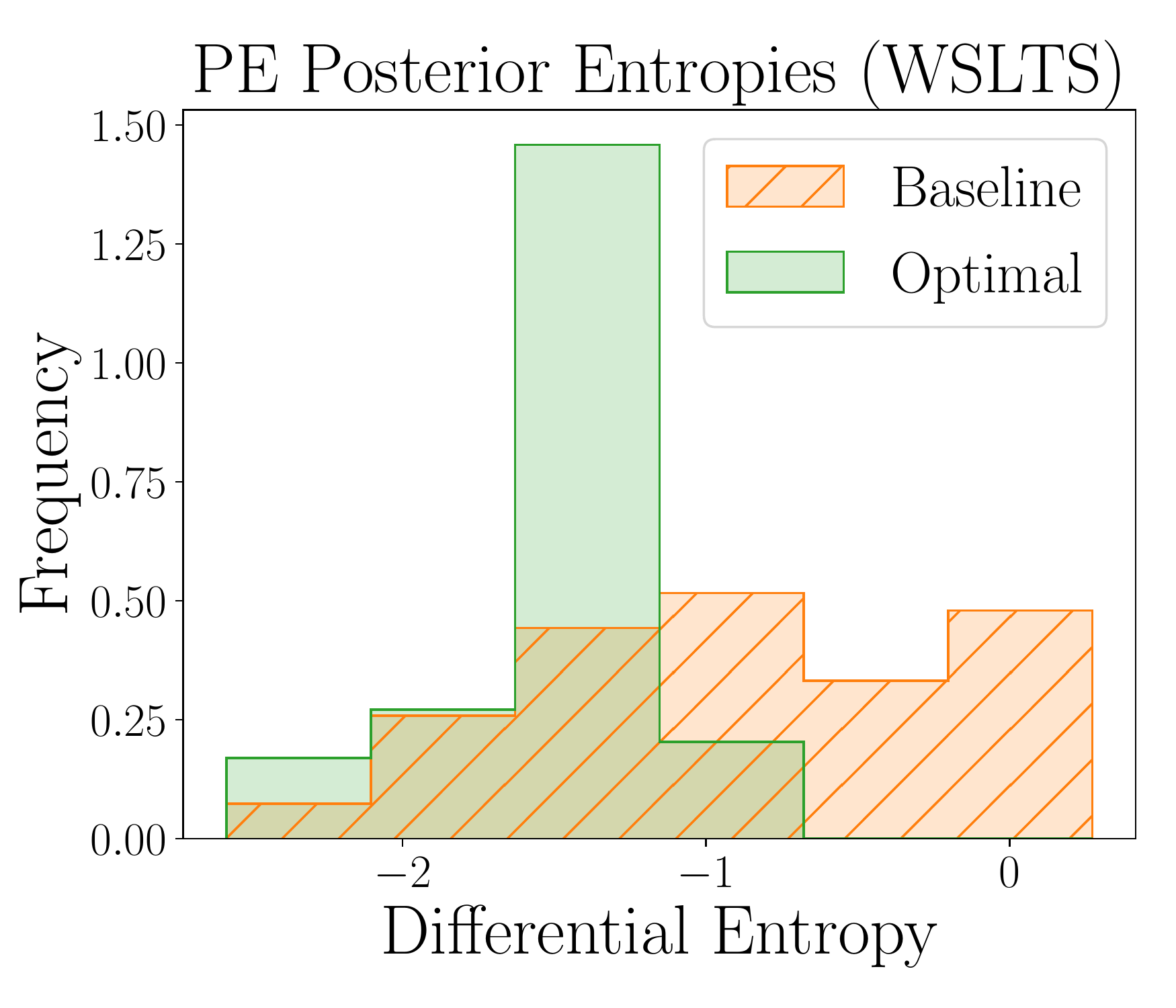}
    \captionof{figure}{Results for the parameter estimation task of the WSLTS model, showing the distribution of posterior differential entropies obtained for optimal (green) and baseline (orange) designs (lower is better).}
    \label{fig:pe_real_entropy_wslts}
\end{minipage}
\hfill
\begin{minipage}{0.48\linewidth}
    \includegraphics[width=\linewidth]{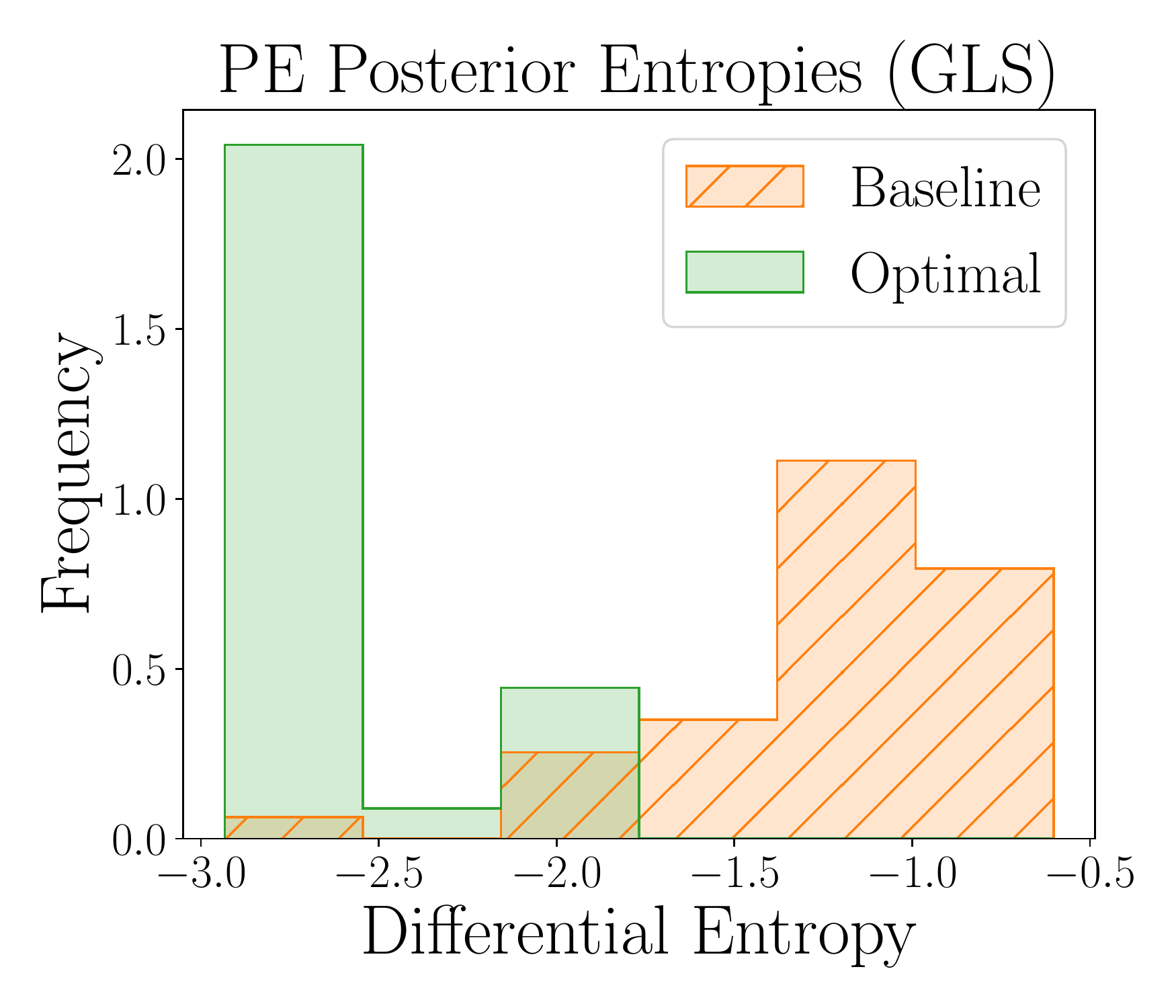}
    \captionof{figure}{Results for the parameter estimation task of the GLS model, showing the distribution of posterior differential entropies obtained for optimal (green) and baseline (orange) designs (lower is better).}
    \label{fig:pe_real_entropy_gls}
\end{minipage}
\end{center}

\begin{center}
    \includegraphics[width=\linewidth]{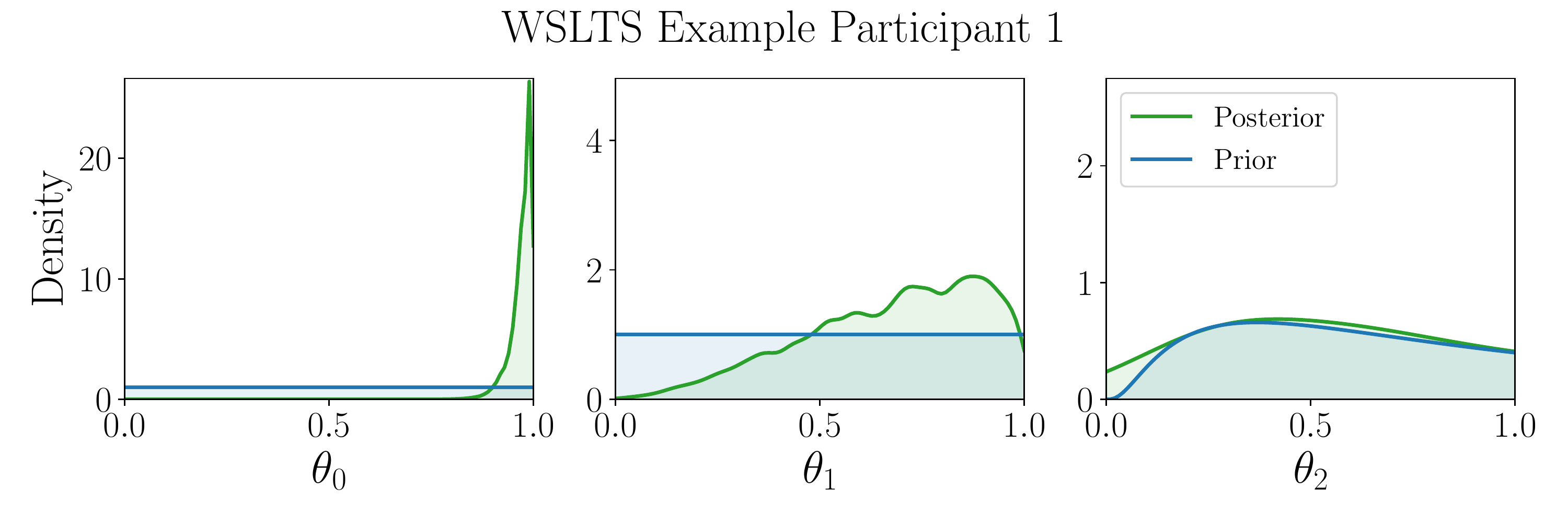}
    \captionof{figure}{Marginal posterior distributions (green) of the WSLTS model parameters for example participant 1.}
    \label{fig:pe_real_post_wslts_ex1}
\end{center}

\begin{center}
    \includegraphics[width=\linewidth]{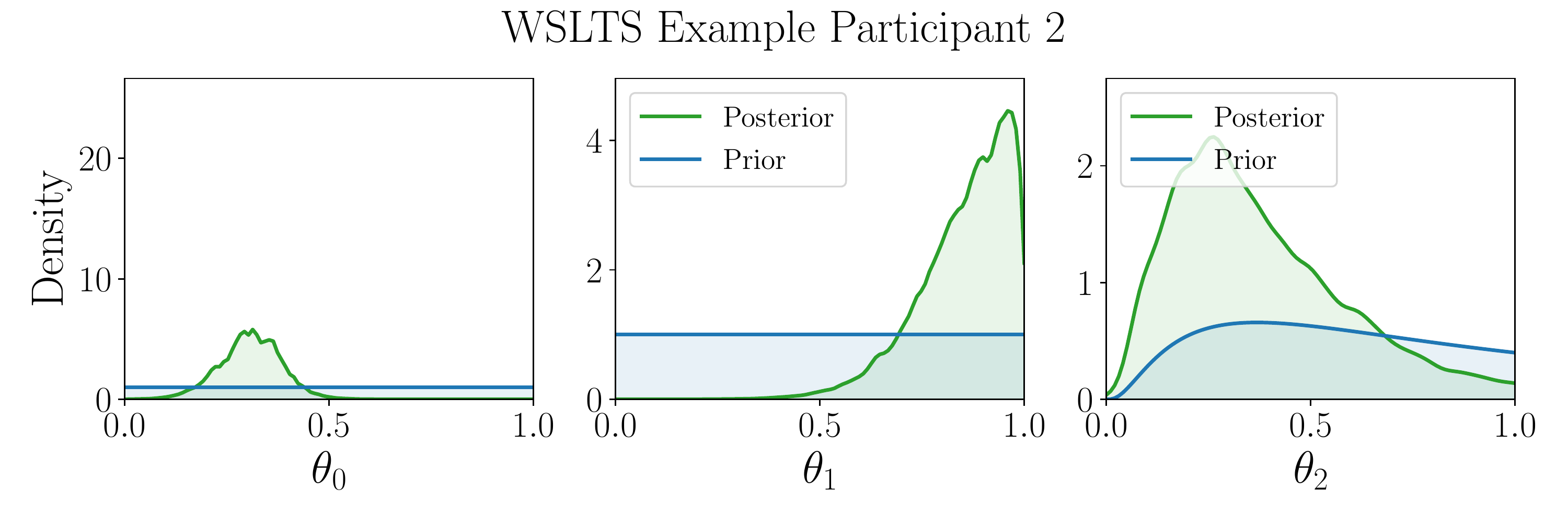}
    \captionof{figure}{Marginal posterior distributions (green) of the WSLTS model parameters for example participant 2.}
    \label{fig:pe_real_post_wslts_ex2}
\end{center}

\begin{center}
    \includegraphics[width=\linewidth]{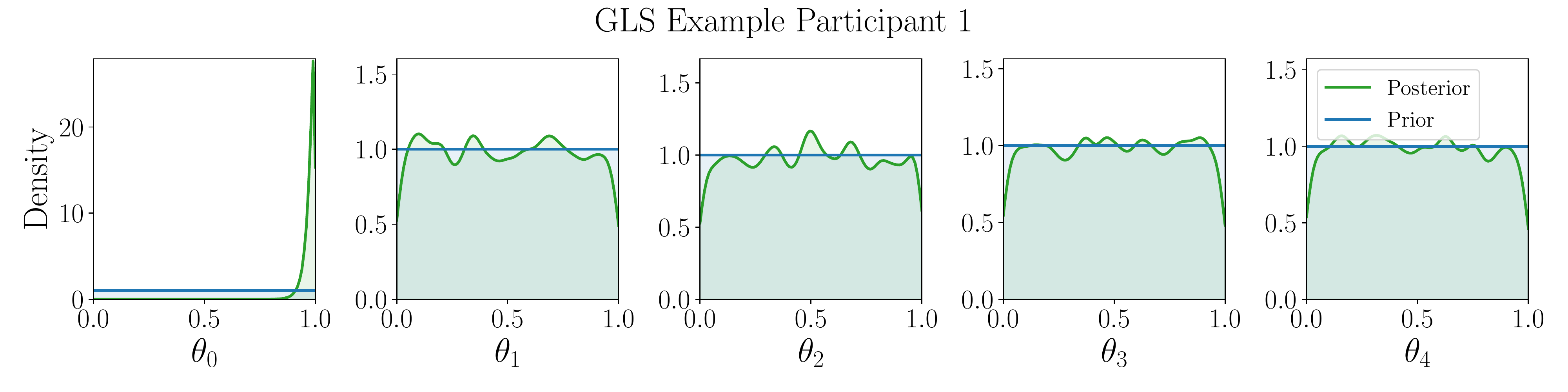}
    \captionof{figure}{Marginal posterior distributions (green) of the GLS model parameters for example participant 1.}
    \label{fig:pe_real_post_gls_ex1}
\end{center}

\begin{center}
    \includegraphics[width=\linewidth]{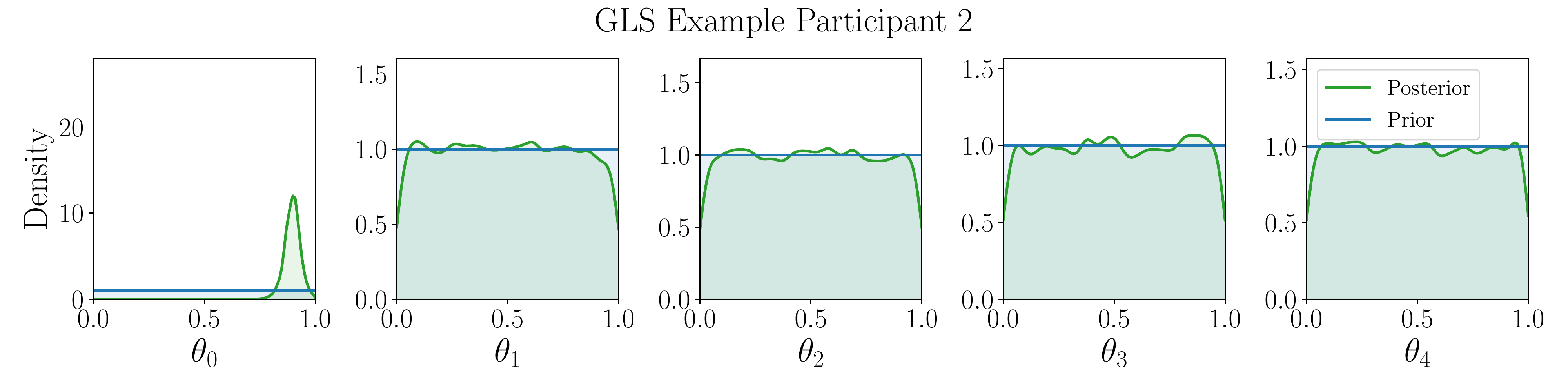}
    \captionof{figure}{Marginal posterior distributions (green) of the GLS model parameters for example participant 2.}
    \label{fig:pe_real_post_gls_ex2}
\end{center}

\end{appendixbox}

\begin{appendixbox}
\section{Exploring the utility surface and locally optimal designs}

In this section, we provide an example of how our framework allows for exploring the surface of the utility function and (locally) optimal designs. To do so, we consider the example of the model discrimination task where the optimal design was found to be around $[0, 0, 0.6]$ for the first bandit arm and $[1, 1, 0]$ for the second bandit arm. Our framework maximizes a probabilistic surrogate model, e.g.~a Gaussian process (GP), over the estimated mutual information (MI) values in order to find optimal designs. It is straightforward to extract the learned GP from the final round of training, which allows us to utilize it for post-training analysis and exploration. Since the design space is 6-dimensional, it is difficult to visualize the entirety of the utility function, provided by the mean function of the learned GP, at once. However, it is possible to explore the utility function by means of slicing. An example of visualizing such a slice is shown in Figure~\ref{fig:utility_slicing}. Furthermore, it is possible to systematically search for local optima of the utility function by running stochastic gradient ascent (SGA) over the mean function of the GP, with several restarts. We have done this for 20 restarts for the above example and summarized the (unique) local optima of the utility function in Table~\ref{tab:local_opt}. Note that instead of using SGA, we could also treat the utility as an unnormalized density and sample designs proportional to this density~\citep[see, e.g.,][]{muller1999}.

\begin{center}
    \includegraphics[width=1
\linewidth]{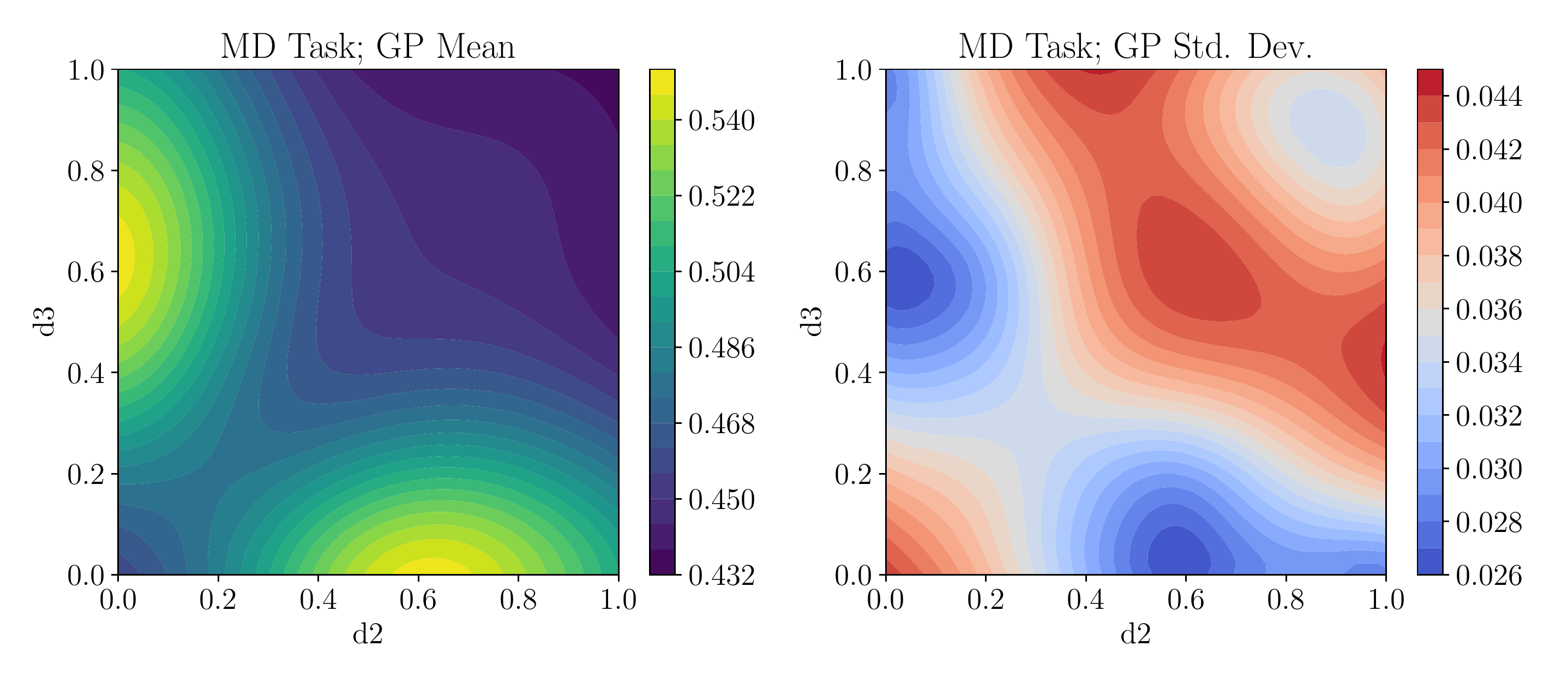}
    \captionof{figure}{Example of how to explore the high-dimensional utility function for the model discrimination (MD) task, by slicing the Gaussian process (GP) learned during the design optimization step. The left plot shows the mean of the GP and the right plot shows the standard deviation of the GP. Shown is the slice corresponding to the design $\mathbf{d} = [[0, d_2, d_3], [1, 1, 0]]$, which contains the global optimum.}
    \label{fig:utility_slicing}
\end{center}

\begin{center}
\begin{tabular}{c | c | cccccc}
\toprule
Rank of Optimum & MI & $d_1$ & $d_2$ & $d_3$ & $d_4$ & $d_5$ & $d_6$ \\
\midrule
1 & 0.672 & 0.000 & 0.000 & 0.607 & 1.000 & 1.000 & 0.000 \\
2 & 0.526 & 1.000 & 0.006 & 1.000 & 0.140 & 0.266 & 0.528 \\
3 & 0.485 & 0.000 & 1.000 & 0.774 & 0.275 & 0.399 & 0.630 \\
4 & 0.479 & 0.683 & 0.775 & 0.000 & 0.216 & 0.717 & 0.128 \\
5 & 0.475 & 0.009 & 0.710 & 0.583 & 0.842 & 0.000 & 0.109 \\
\bottomrule
\end{tabular}
\captionof{table}{A ranking of local design optima for the model discrimination task. Shown are the rank, the mutual information (MI) estimate computed via the Gaussian process (GP) mean, and the locally optimal designs. Note that, for this task, the designs for the first bandit arm were $[d_1, d_2, d_3]$ and $[d_4, d_5, d_6]$ for the second bandit arm. The $5$ unique optima were obtained by running stochastic gradient ascent on the GP mean with $20$ restarts.}
\end{center}
\label{tab:local_opt}
\end{appendixbox}

\begin{appendixbox}
\section{Example participants}

\begin{center}
    \includegraphics[width=0.5\linewidth]{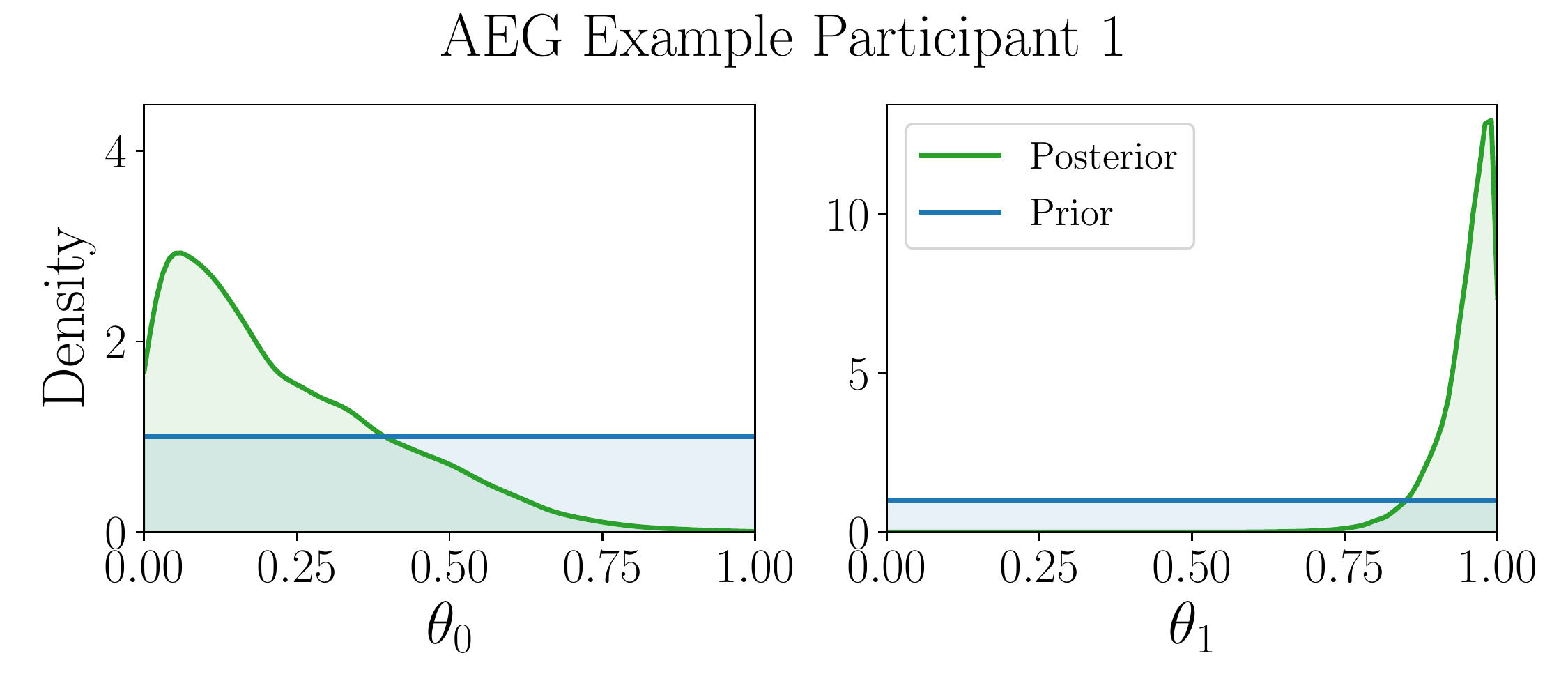}
    \label{fig:example_post1}
    \captionof{figure}{Inferred marginal posterior model parameters for human participants best described by the AEG model. Marginal posterior distributions (green) of the AEG model parameters for example participant 1.}
\end{center}

\begin{center}
    \includegraphics[width=0.5\linewidth]{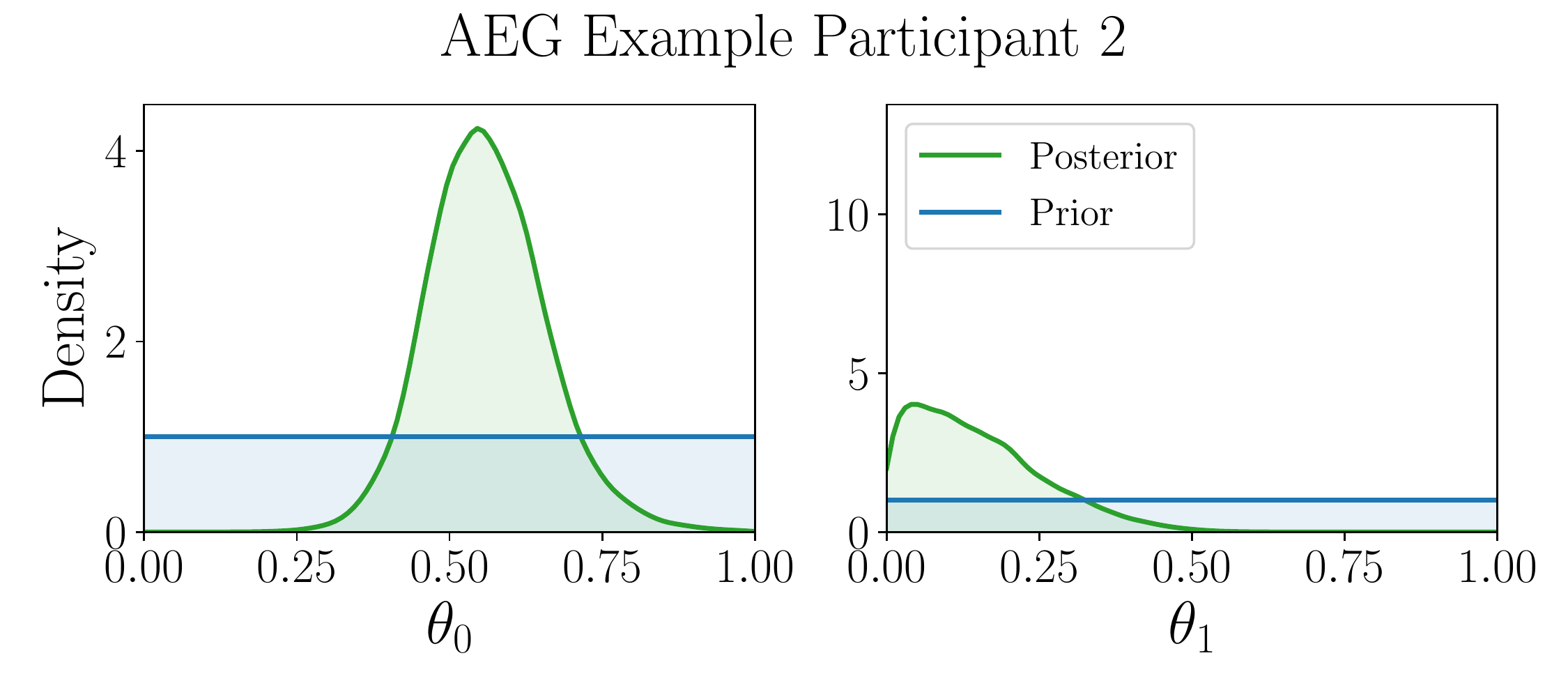}
    \label{fig:example_post2}
    \captionof{figure}{Posterior distribution (green) of the AEG model parameters for example participant 2.}
\end{center}

\begin{center}
    \includegraphics[width=0.5\linewidth]{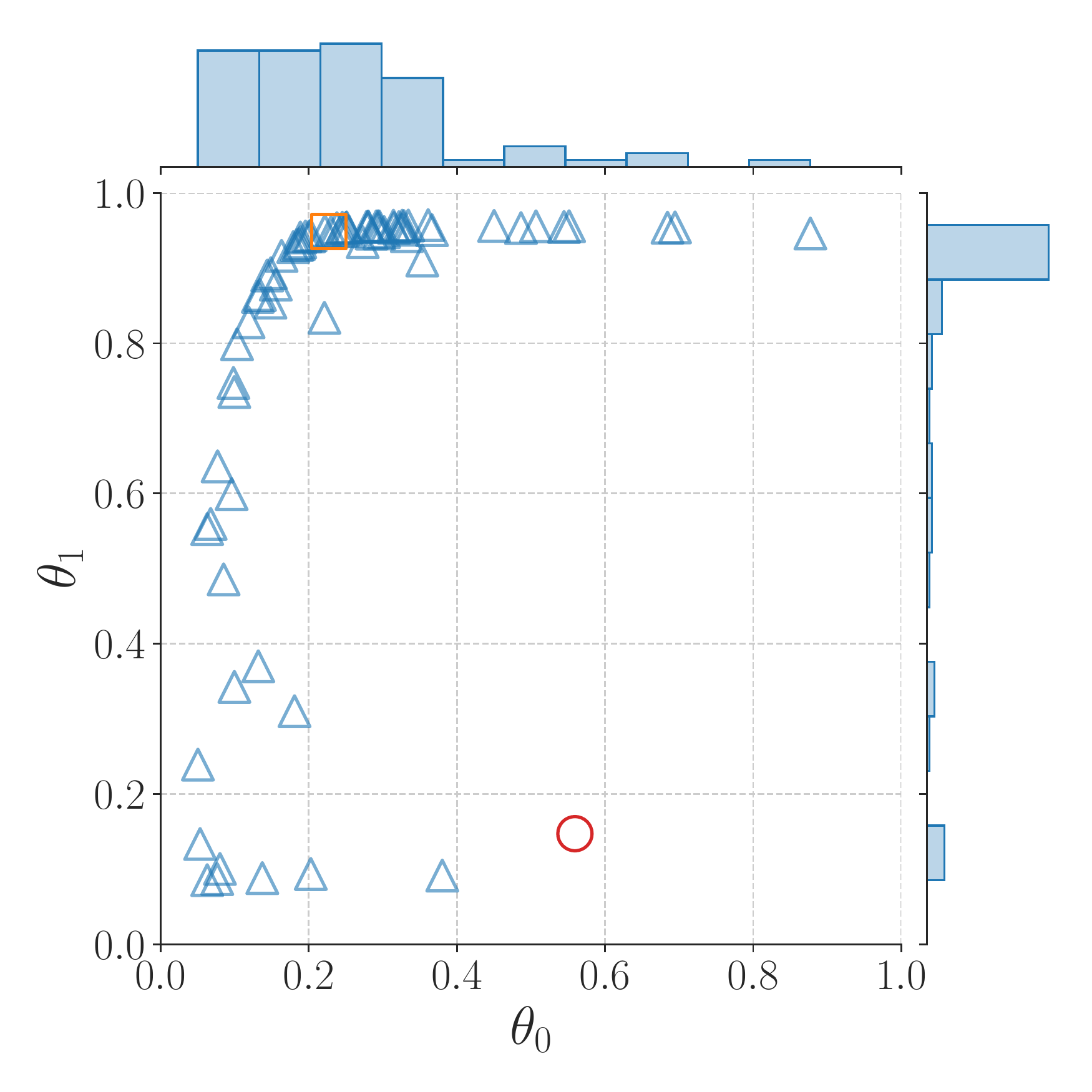}
    \label{fig:scatter}
    \captionof{figure}{Mean values of the parameter posterior distributions for each of the $75$ participants best described by the AEG model; the square and circle markers represent example participants $1$ and $2$, respectively.}
\end{center}

We now explore the data from our optimal design group in the human participant study, specifically focusing on participants that were best described by the AEG model as an example. We qualitatively investigate their behavior by visualizing the (marginal) posterior distribution of the AEG model, as shown in Figures~\ref{fig:example_post1} and~\ref{fig:example_post2} and  for two example participants. The posterior distribution for the first participant (Figure~\ref{fig:example_post1}) indicates small values for the first parameter $\theta_0$ of the AEG model, i.e.~the $\varepsilon$ parameter, which implies that the participant tends to select the arm with the highest probability of receiving a reward as opposed to making random exploration decisions. Larger values of the second parameter $\theta_1$ of the AEG model, i.e.~the ``stickiness'' parameter, imply that the participant is biased towards re-selecting the previously chosen bandit arm. The second example participant (Figure~\ref{fig:example_post2}) shows markedly different behavior. As indicated by the marginal posterior of the $\theta_0$ parameter, this participant engages in considerably more random choices. Moreover, this participant shows ``anti-sticky'' behavior, as implied by low values of the second parameter $\theta_1$.
Lastly, we compute the posterior mean for each participant best described by the AEG model and visualize them in a scatter plot in Figure~\ref{fig:scatter}. We find that most participants best described by the AEG model show behavior somewhat similar to the first example participant, but as expected, there is inter-individual variation, e.g., with some participants displaying ``anti-sticky'' behavior.

\end{appendixbox}
\end{document}